\documentclass[times, review, 10pt]{elsarticle}
\usepackage{lineno,hyperref}
\usepackage{array}
\usepackage{amssymb}
\usepackage{enumerate}
\usepackage{makecell}
\usepackage{mathtools}
\usepackage{array}
\usepackage{multirow}
\usepackage{ulem}
\usepackage{color}
\usepackage{diagbox}
\usepackage{enumerate}
\usepackage{graphicx}
\usepackage{threeparttable}
\usepackage{subfigure}
 \usepackage{tikz,mathpazo} 
\usepackage{indentfirst}
\usepackage{soul}
\usepackage{amsmath}
\usepackage[justification=centering]{caption}
\usepackage{tabularx}
\usepackage{algorithm}
\usepackage{algorithmicx}
\usepackage{algpseudocode}
\usepackage{adjustbox}
\usepackage{geometry}
\usepackage{arydshln}
\usetikzlibrary{shapes.geometric, arrows}
\modulolinenumbers[5]

\soulregister{\cite}7
\soulregister{\ref}7

\newcommand{\circledsmall}[1]{\lower.7ex\hbox{\tikz\draw (0pt, 0pt)%
    circle (.5em) node {\makebox[0.1em][c]{\small#1}};}}

\newcommand{\circledtiny}[1]{\lower.7ex\hbox{\tikz\draw (0pt, 0pt)%
    circle (.3em) node {\makebox[0.1em][c]{\tiny #1}};}}

\journal{Journal of \LaTeX\ Templates}

\begin{document}

\begin{frontmatter}

\title{Co-Evidential Fusion with Information Volume for Medical Image Segmentation}

    \author[label1,label2]{Yuanpeng He\corref{cor1}}
    \author[label3]{Lijian Li}
    \author[label3]{Tianxiang Zhan}
    \author[label3]{Chi-Man Pun}
    \author[label1,label2]{Wenpin Jiao}
    \author[label1,label2]{Zhi Jin}
    
    \affiliation[label1]{organization={Key Laboratory of High Confidence Software Technologies (Peking University), Ministry of Education},
    	city={Beijing},
    	postcode={100871},
    	country={China}}
    
    \affiliation[label2]{organization={School of Computer Science, Peking University},
    	city={Beijing},
    	postcode={100871},
    	country={China}}

    \affiliation[label3]{organization={Department of Computer and Information Science, University of Macau},
    	city={Macau},
    	postcode={999078},
    	country={China}}

    \cortext[cor1]{Corresponding author: Yuanpeng He is with Key Laboratory of High Confidence Software Technologies (Peking University), Ministry of Education, Beijing, China; School of Computer Science, Peking University, Beijing, China. E-mail address: heyuanpeng@stu.pku.edu.cn and heyuanpengpku@gmail.com}

\begin{abstract}
Although existing semi-supervised image segmentation methods have achieved good performance, they cannot effectively utilize multiple sources of voxel-level uncertainty for targeted learning. Therefore, we propose two main improvements. First, we introduce a novel pignistic co-evidential fusion strategy using generalized evidential deep learning, extended by traditional D-S evidence theory, to obtain a more precise uncertainty measure for each voxel in medical samples. This assists the model in learning mixed labeled information and establishing semantic associations between labeled and unlabeled data. Second, we introduce the concept of information volume of mass function (IVUM) to evaluate the constructed evidence, implementing two evidential learning schemes. One optimizes evidential deep learning by combining the information volume of the mass function with original uncertainty measures. The other integrates the learning pattern based on the co-evidential fusion strategy, using IVUM to design a new optimization objective. Experiments on four datasets demonstrate the competitive performance of our method.
\end{abstract}

\end{frontmatter}
\section{Introduction}

In the contemporary medical domain, image segmentation techniques have become an indispensable computational tool, playing a pivotal role in clinical diagnosis, treatment planning, and biomedical research. Medical image segmentation refers to the process in which algorithms accurately delineate anatomical structures, pathological lesions, or other regions of interest from complex medical imagery, thereby enabling precise interpretation and analysis of physiological conditions \cite{tragakis2023fully, li2023lvit}. This technique significantly enhances diagnostic precision, facilitates the design of personalized therapeutic strategies, and supports longitudinal disease monitoring. In particular, accurate delineation of tumors can directly inform surgical and radiological interventions, tailoring procedures to individual patients and improving clinical outcomes.

Driven by advancements in deep learning \cite{xu2023spatio}, medical image segmentation has witnessed transformative growth in recent years \cite{DBLP:conf/aaai/FengJWSZG23, DBLP:conf/icml/0001RYCK23, balaji2023image}. Convolutional Neural Networks (CNNs) are particularly effective in learning hierarchical spatial features, while Variational Autoencoders (VAEs) provide robust probabilistic frameworks for both generative modeling and representation learning. These approaches, despite their success, are predominantly supervised, requiring extensive labeled datasets—a luxury rarely afforded in clinical scenarios due to high annotation cost, expertise demand, and time constraints.

To mitigate this limitation, semi-supervised learning (SSL) strategies have emerged as a promising solution. Techniques such as self-training \cite{chaitanya2023local}, generative adversarial networks (GANs) \cite{zhang2023multi}, and graph-based models \cite{meng2022dual} leverage both labeled and unlabeled data to enhance model generalization. However, SSL methods face critical challenges: error propagation in self-training \cite{jiang2023review}, instability in GAN training \cite{abdusalomov2023evaluating}, and difficulty in managing high uncertainty and fuzzy boundaries in graph-based techniques. Furthermore, the heterogeneity of medical images—spanning modalities (CT, MRI, Ultrasound), device differences, and pathological variability—further exacerbates the challenge of building robust segmentation systems \cite{wu2023exploring, qiu2023federated, wang2023mcf, rangnekar2023semantic}.

To address these limitations, we propose a novel approach that integrates generalized evidential learning based on Dempster-Shafer (D-S) theory and co-evidential fusion strategies within a semi-supervised segmentation framework. Our method builds upon a growing body of work that explores uncertainty modeling and evidential reasoning in complex decision-making tasks \cite{he2021conflicting, he2022new, he2022mmget, he2022ordinal, he2023ordinal, he2023tdqmf}. Specifically, we introduce a pignistic evidential fusion mechanism to robustly combine outputs from heterogeneous classifiers, enabling the model to capture uncertainty signals and suppress noisy pseudo-labels in unlabeled data \cite{he2024uncertainty, he2024epl, he2025co}. Key to our framework is the Information Volume of Mass Function (IVUM), which provides a principled measure of belief strength and uncertainty, facilitating informed optimization of fusion strategies \cite{he2024epl, he2025co}. This allows the segmentation model to self-assess the credibility of predictions and adaptively refine representations for ambiguous regions—particularly beneficial in scenarios involving limited annotated samples. Experimental validation across multiple datasets demonstrates the effectiveness of this method in reducing dependency on labeled data while maintaining high segmentation accuracy.

Our contributions further extend to designing multi-head attention mechanisms and hybrid evidential networks for weak-supervised tasks \cite{he2024generalized}, as well as integrating evidential reasoning into residual feature reuse architectures \cite{he2024residual}, and exploring prototype refinement strategies through multi-prototype learning \cite{bi2025multi, li2024efficient}. In addition, we incorporate domain knowledge through quantum mass function modeling \cite{he2023tdqmf}, Pythagorean fuzzy matrices \cite{he2021matrix}, and develop advanced federated frameworks for hospital-level collaboration and semi-supervised segmentation \cite{huang2025unitrans, he2024mutual}. Furthermore, our efforts are synergistic with parallel developments in aspect-based sentiment analysis \cite{li2022nndf}, time series forecasting \cite{zhan2023differential, zhan2024time}, knowledge graph completion \cite{li2025rethinking, li2025towards}, and code generation with self-debugging frameworks \cite{chen2025revisit}. These collectively reflect the versatility of evidential modeling and entropy-guided learning across domains \cite{li2025adaptive}. 

In sum, the main contributions of this paper are summarized as follows:
\begin{itemize}
	\item We propose the strategy of co-evidential fusion to use the original annotated data to further assist the model in learning the mixture of labeled features and the relationship between labeled and unlabeled data.
	\item We introduce the concept of information volume of mass function into evidential deep learning and extend its framework through bringing the traditional D-S evidence theory so as to provide a more precise assessment of the evidences in different dimensions.
	\item We propose novel evidential optimization objectives with constructed evidences under the generalized framework of evidential deep learning and one of the learning objectives depends on the previously designed learning modes, which enable model to carry out multi-level learning and obtain a better understanding of features.
	\item The experiments on four datasets with three different ratios of labeled data demonstrate the superior performance of the method proposed in this paper.
\end{itemize}

The rest of this paper is organized as follows: Section \ref{sec:rw} presents a review of related literature. Section \ref{sec:met} introduces the proposed methods. A detailed experimental analysis is presented in Section \ref{sec:exp}. Finally, we draw the conclusion in Section \ref{sec:con}.

\section{Uncertainty-Based Semi-Supervised Medical Image Segmentation}\label{sec:rw}
In recent field of uncertainty-based methods, significant progress has been made for semi-supervised medical image segmentation. A common feature of these studies is the use of the concept of uncertainty to enhance the performance of semi-supervised learning, especially in the case of limited labeled data. Uncertainty plays a central role in dealing with noisy and uncertain data in semi-supervised learning. For example, Wang et al. \cite{wang2022uncertainty} proposes an uncertainty-guided pixel-contrast-based learning method, enabling the model to be trained more efficiently with unlabeled data by constructing uncertainty mappings and reducing noise sampling. In addition, the uncertainty-aware pseudo-labeling and consistency approach \cite{lu2023uncertainty} avoids the limitation of fixed threshold screening for pseudo-labeling by directly utilizing uncertainty to correct the learning of noisy pseudo-labeling. For the design of learning frameworks, several studies have adopted dual-view or multi-task learning strategies. The uncertainty-guided dual-view framework \cite{peiris2023uncertainty} and the triple uncertainty-guided average teacher model \cite{wang2022semi} emphasize the importance of extracting more reliable knowledge from different perspectives or tasks. These approaches utilize uncertainty as an indicator to guide the learning process, thereby increasing the efficiency of the model in utilizing unlabeled data. Furthermore, studies have shown that the estimation and application of uncertainty not only improve the performance of semi-supervised medical image segmentation, but also enhance the robustness and generalization of the model. The inconsistency-aware uncertainty estimation method \cite{DBLP:journals/tmi/Shi0LL0Y0022} and the concept of self-recursive uncertainty \cite{10.1007/978-3-030-59710-8_60} have demonstrated the potential of uncertainty in improving the effectiveness of semi-supervised learning. In summary, uncertainty-based semi-supervised medical image segmentation methods effectively improve the learning ability and performance of models with limited labeled data by integrating uncertainty concepts and strategies.

\section{Methodology}\label{sec:met}
With respect to the task of semi-supervised medical image segmentation, the definition of 3D volume of medical image can be given as $\mathbf{X} \in \mathbb{R}^{W\times H \times L}$. The purpose of semi-supervised medical segmentation is to predict the label map $\textbf{Y} \in \{0,1,...,S - 1\}^{W\times H \times L}$ for each voxel which indicates where the backgrounds and targets are located in $\mathbf{X}$. Specifically, $S$ represents the number of classes. And the corresponding data set $\zeta$ which is utilized for training is composed of $Q$ labeled data $\zeta^{l}$ and $W$ unlabeled data $\zeta^{u}$ in which $W$ is much larger than $Q$. In the pre-training stage, the model is trained utilizing labeled data which consists of two parts $\zeta^{l} = \zeta^{l_1} \cup \zeta^{l_2}$ where $\zeta^{l_1} = \{\textbf{X}_{a}^{l_1}, \textbf{Y}_{a}^{l_1}\}_{a=1}^{\lfloor Q / 2 \rfloor}$ and $\zeta^{l_2} = \{\textbf{X}_{b}^{l_2}, \textbf{Y}_{b}^{l_2}\}_{b = \lfloor Q / 2 \rfloor + 1}^{Q}$. And for the self-training stage, the model is supposed to train on both labeled and unlabeled data which can be given as $\zeta = \zeta^{l} \cup \zeta^{u}$ where $\zeta^{l} = \{\textbf{X}_{c}^{l}, \textbf{Y}_{c}^{l}\}_{c=1}^{Q}$ and $\zeta^{u} = \{\textbf{X}_{c}^{u}\}_{c=Q+1}^{Q+W}$.

And the data pre-processing strategies differ in the pre-training and self-training stages. With respect to the pre-training stage, two labeled images  $\{\textbf{X}_{a}^{l_1},\textbf{X}_{b}^{l_2}\}$ are randomly picked from $\zeta^{l_1}$ and $\zeta^{l_2}$ respectively. A crop is copy-pasted from $\textbf{X}_{a}^{l_1}$ to $\textbf{X}_{b}^{l_2}$ to produce a new image $\textbf{X}^{\mathbb{M}_{1}}$ using mask $\mathcal{M}_1$. Correspondingly, the same adjustment is made on the labels of two images to generate a new label $\textbf{Y}^{\mathbb{M}_{1}}$. The newly produced images and labels are fed into one model to obtain a set of initialization parameters $\mathcal{I}_{s1}$. Instead, another mask $\mathcal{M}_2$ is used to do the copy-paste between two sets of original labeled data for acquiring another set of initialization parameters $\mathcal{I}_{s2}$. And $\mathcal{I}_{s1}$ and $\mathcal{I}_{s2}$ are utilized for the two kinds of sub-networks, respectively. More specifically, in addition to the use of mixed images and labels for supervised training, we also feed two sets of original labeled data into the model for training. We convert the predictions obtained from these three training processes into evidences respectively, and the latter two are fused with the evidence corresponding to the mixed data respectively, so that the model can further learn from the features in the original information while learning the semantic features of the mixed data. For the self-training stage, different from the operation of pre-training stage, the copy-paste operation is carried out on the labeled and unlabeled data. In detail, assume there exist two labeled images $\{\textbf{X}_{i}^{l}, \textbf{X}_{j}^{l}\}$ and two unlabeled images $\{\textbf{X}_{n}^{u}, \textbf{X}_{m}^{u}\}$ which are randomly picked. Then, a random crop is copy-pasted from $\textbf{X}_{i}^{l}$ (foreground) to $\textbf{X}_{m}^{u}$ (background) and from $\textbf{X}_{n}^{u}$ (foreground) to $\textbf{X}_{j}^{l}$ (background) to generate mixed images $\textbf{X}_{lu}^{\mathbb{M}_1}$ and $\textbf{X}_{lu}^{\mathbb{M}_2}$. The operation is designed to establish the semantic relationship between the labeled data and the unlabeled data, so that the model can learn to obtain the ability to make predictions on the unlabeled data. Specifically, $\textbf{X}_{lu}^{\mathbb{M}_1}$ and $\textbf{X}_{lu}^{\mathbb{M}_2}$ are fed into the model to produce corresponding segmentation masks for $\textbf{Y}^{\mathbb{M}_{1}}_{lu}$ and $\textbf{Y}^{\mathbb{M}_{2}}_{lu}$. Similar to the training strategy in the pre-training stage, we also feed the labeled data into the model to generate supervised predictions which are transformed into evidences. The evidences from supervised predictions are expected to be fused with the ones from mixed images, which aims to assist the model in learning the semantic features of mixed images referencing supervised signals. Besides, for the optimization of each network, it is supposed to utilize one of the sub-networks to generate predictions of unlabeled images for another sub-netowrk which is utilized to supervise segmentation masks.

\subsection{Preliminaries}

\subsubsection{Evidential Deep Learning}
In recent years, with the rapid development and wide application of deep learning, how to effectively assess and express model uncertainty has become a pressing issue. Evidential Deep Learning (EDL) \cite{DBLP:conf/acl/WuZW23} is an emerging approach in the field of deep learning that aims to quantify and express uncertainty by introducing evidence theory \cite{DBLP:series/sfsc/Yager08}. The core idea of this approach is to utilize evidence theory ($e.g.$, Dempster-Shafer theory) and probability distributions ($e.g.$, Dirichlet distribution) to express and process uncertainty information. In traditional deep learning models, uncertainty is usually ignored or expressed only indirectly through the probability distribution of the model output. However, this approach often does not accurately reflect how confident the model is about its predictions. In contrast, EDL provides a more direct and flexible way to express uncertainty not only in the predictions, but also in the data itself.
Sensoy et al. \cite{sensoy2018evidential} treats the predictions of a neural network as subjective views by applying the Dirichlet distribution to the category probabilities and learning to collect evidence for these views from the data. This approach achieves significant results in detecting out-of-distribution queries and resistance to adversarial perturbations, demonstrating the potential of EDL to improve model robustness.
Amini et al. \cite{amini2020deep} explores how to train non-Bayesian neural networks to estimate continuous targets and their associated evidence. This approach offers significant advantages in terms of efficiency and scalability, enables efficient uncertainty learning without relying on sampling, and is robust to adversarial and out-of-distribution samples.
Han et al. \cite{DBLP:journals/pami/HanZFZ23} utilizes multiple perspectives to promote classification reliability and robustness, and provided a unified learning framework for models to accurately estimate uncertainty by applying the Dirichlet distribution and Dempster-Shafer theory.
Deng et al. \cite{10.5555/3618408.3618708} proposes an EDL method based on Fisher information. This approach shows significant advantages when solving samples with high uncertainty, especially in small sample classification scenarios.
In summary, evidential deep learning represents important advances in the field of deep learning for uncertainty modeling and estimation. These works provide new tools and ideas for deep learning modeling, enrich the understanding of uncertainty, and are expected to improve the reliability and robustness of models in various applications. 

More specifically, EDL is developed on the basis of subjective logic-based evidence theory \cite{DBLP:books/sp/Josang16}. For a given prediction task with $N$ classes, assume there exists a vector of prediction for a sample $\psi$, $v \in \mathbb{R}^{N}_{+}$. Then, the corresponding Dirichlet distribution $\mathcal{D}_d$ can be defined as:
\begin{equation}
	\mathcal{D}_d(\mathbf{t}|\mathbf{c}) = \left\{
	\begin{aligned}
		\frac{1}{\mathbb{D}(\mathbf{c})}\prod_{k=1}^{N}t_k^{c_k -1}, for\ \mathbf{t} \in \mathbb{S}_N \\
		0, \ otherwise \qquad \\
	\end{aligned}
	\right.
\end{equation}
where $\varphi_k = v_k+1, k = 1,...,N$. $\mathbb{D}$ represents an $N$-dimensional beta function and $\mathbf{t}$ denotes a point on the $N$-dimensional unit simplex $\mathbb{S}_N$. Every point of the simplex, is regarded as a point estimate of the probability distribution. Treating $\mathcal{D}_d(\mathbf{t}|\omega)$ as a prior on the likelihood function $\mathbb{L}(\mathbf{y}|\mathbf{t})$, the corresponding marginal likelihood function $\mathbb{M}(\mathbf{y}|\varphi)$ can be obtained:
\begin{equation}
	\begin{aligned}
		\mathbb{M}(\mathbf{y}|\mathbf{c})& = \int\mathbb{L}(\mathbf{y}|\mathbf{t})\mathcal{D}_d(\mathbf{t}|\mathbf{c})\\&=\int_{\mathbf{t} \in \mathbb{S}_N}\prod_{k=1}^{N}t_k^{y_k}\frac{1}{\mathbb{D}(\mathbf{c})}\prod_{k=1}^{N}t_k^{c_k -1}d\mathbf{t}
	\end{aligned}
\end{equation}
where $\mathbf{y}$ denotes one-hot ground-truth vector for sample $\psi$. Based on the definition of $\mathbb{M}(\mathbf{y}|\varphi)$, the traditional and most approbatory optimization target of vanilla EDL, such as the negative logarithm of the marginal likelihood, which can be defined as:
\begin{equation}
	\mathcal{L}_{EDL} = -log(\mathbb{M}(\mathbf{y}|\mathbf{c})) = \sum_{k=1}^{N}y_k(log \mathcal{U}-log c_k)
\end{equation}

And for the two main constituent elements in evidential deep learning, namely uncertainty measure of evidence $\mathcal{U}$ and probability of prediction for classes, $\mathcal{U}$ and $\widetilde{p}$, which can be defined as:
\begin{equation}
	\mathcal{U} = N/T, \quad \widetilde{p}_k = v_k/T
\end{equation}
where $T = \sum_{k=1}^{N}c_k$. Obviously, when the belief values become bigger, the degree of uncertainty gets smaller. It can be concluded that the belief values are inversely proportional to the uncertainty measure.
\subsubsection{D-S Evidence Theory}
D-S evidence theory is also called Dempster-Shafer evidence theory which was proposed by Dempster \cite{DBLP:series/sfsc/Dempster08a} and further developed by Shafer \cite{DBLP:journals/ijar/Shafer16}. D-S evidence theory is more flexible than classical Bayesian probability and able to model uncertainty directly. D-S evidence theory plays a pivotal role across diverse disciplines by addressing uncertainty and fusing heterogeneous evidence, particularly excelling in fields requiring robust decision-making under incomplete information. In robotics and autonomous systems, D-S evidence theory enhances sensor fusion reliability by resolving conflicts between LiDAR, camera, and radar data while explicitly modeling ignorance, a critical advantage over probabilistic methods. Cybersecurity applications leverage D-S evidence theory to aggregate low-confidence intrusion alerts into high-confidence threat assessments, reducing false positives. Environmental science employs D-S evidence theory for risk modeling under overlapping climate scenarios, enabling adaptive policy decisions. In semi-supervised medical domains, where labeled data scarcity and diagnostic uncertainty prevail, D-S evidence theory significantly enhances performance by quantifying epistemic uncertainty through belief functions, allowing models to distinguish between "lack of knowledge" and probabilistic randomness. It improves pseudo-labeling reliability by fusing evidence from multiple weakly supervised models or unlabeled data clusters, mitigating noise propagation. For instance, in medical image analysis, D-S evidence theory-based fusion of radiomic features and clinical data refines tumor classification, while its ability to represent partial ignorance aids in handling ambiguous diagnoses (e.g., early-stage lesions). By integrating D-S evidence theory with deep semi-supervised frameworks, medical systems achieve higher robustness against class imbalance and annotation sparsity, as belief assignments guide adaptive sample weighting and uncertainty-aware consistency regularization, ultimately enhancing diagnostic accuracy and model trustworthiness in resource-constrained clinical settings. To be specific, assume there exists a non-empty set $\mathcal{S}$ which includes $N$ mutually exclusive propositions. On the basis of definition of $\mathcal{S}$, the framework of discernment (FOD) can be defined as:
\begin{equation}
	\mathcal{S} = \{P_1, ..., P_N\}
\end{equation}

With the concept of FOD, the power set of $\mathcal{S}$ which contains $2^{|\mathcal{S}|}$ can be given as:
\begin{equation}
	2^{\mathcal{S}} = \{\{\emptyset\},\{P_1\},...,\{P_1,P_2\},...,\{\mathcal{S}\}\}
\end{equation}
where $|\mathcal{S}|$ indicates the cardinality of $\mathcal{S}$. On the basis of the definition of power set $2^{\mathcal{S}}$, the corresponding basic probability assignment which is also called mass function can be defined as:
\begin{equation}
	\varepsilon: 2^{\mathcal{S}} \rightarrow [0,1]
\end{equation} 

More specifically, two main properties which basic probability assignment is supposed to satisfy can be given as:
\begin{equation}
	\varepsilon(\emptyset) = 0,\quad \sum_{p \subseteq 2^{|\mathcal{S}|}}\varepsilon(p) = 1
\end{equation}
where $\sigma(p)$ represents the degree of support for the proposition $p$ when the aforesaid properties can be satisfied. Besides, for the fusion of $\mathbb{F}$ given evidences, the rule of combination can be defined as:
\begin{equation}\small
	\varepsilon_{fused}(A) = \frac{1}{1-\mathcal{K}} * \sum_{G_d \cap 
		H_f \cap ... \cap J_g = A} \varepsilon_1(G_d) * \varepsilon_2(H_f) * ... * \varepsilon_{\mathbb{F}}(J_g)
\end{equation}
where $\varepsilon_{fused}(A)$ represents the final belief degree of proposition $A$. $\mathcal{K}$ denotes the conlicting degree among evidences which can be defiend as:
\begin{equation}
	\mathcal{K} = \sum_{G_d \cap 
		H_f \cap ... \cap J_g = \emptyset} \varepsilon_1(G_d) * \varepsilon_2(H_f) * ... * \varepsilon_{\mathbb{F}}(J_g)
\end{equation}

Besides, some researchers seek for a more effective method for better utilization of uncertainty measure, the pignistic transformation is proposed to adapt to the framework of D-S evidence theory accordingly \cite{DBLP:conf/uai/Smets89}. Assume there exists a proposition $\delta$, the corresponding pignistic transformation can be defined as:
\begin{equation}
	BeT_\varepsilon(\delta) = \sum_{\varrho \subseteq \mathcal{S}, \delta \in \varrho}\frac{\varepsilon(\varrho)}{|\varrho|}
\end{equation}

\subsubsection{Information Volume of Mass Function}
In information theory \cite{DBLP:journals/bstj/Shannon48}, entropy is usually utilized to measure the degree of uncertainty with respect to a system. In the field of D-S evidence theory, a novel entropy named Deng entropy \cite{DBLP:journals/chinaf/Deng20} is proposed to describe uncertainty contained in mass functions. However, given a probability distribution, its corresponding information volume can be given by Shannon entropy, but how to calculate information volume for a mass function is still an open issue. Recently, a kind of information volume measure based on Deng entropy is proposed \cite{deng2020information}. Assume there exists a mass function, the corresponding frame of discernment can be given as $\mathcal{S} = \{P_1, ..., P_N\}$. Specifically, index $i$ is utilized to represent the times of calculation loop, and $m(A_i)$ denotes separate mass function of different loops. Then, the information volume of mass function can be obtained using the following steps:

\begin{description}
	\item[\textbf{Step 1:}]\ \ Input mass function $m(A_0)$.
	\item[\textbf{Step 2:}]\ \ If the mass function of the element whose cardinality is larger than 1, then it is supposed to be continuously separated until convergence. Concretely, the loop is repeated in the following three sub-steps until Deng entropy is convergent.

	\item \textbf{1)} For an element whose cardinality is greater than 1, its mass function is expected to be separated according to the proportion produced by the maximum Deng entropy:
	\begin{equation}
		\varepsilon(A_i) = \frac{(2^{|A_i|}-1)}{\sum_{A_i \in 2^{\mathcal{S}}}(2^{|A_i|}-1)}
	\end{equation}
	\item \textbf{2)} With the definition of Deng entropy, uncertainty of all the mass functions is supposed to be calculated excluding those who have already been divided. And the corresponding result can be denoted as $E_i(\varepsilon)$.
	\item \textbf{3)} Calculate $\Delta_i = E_i(\varepsilon) - E_{i-1}(\varepsilon)$. When $\Delta_i$ meets the following restriction, the calculation loop is jumped out:
	\begin{equation}
		\Delta_i = E_i(\varepsilon) - E_{i-1} < \rho
	\end{equation}
	where $\rho$ represents an allowable error.
	
	\item[\textbf{Step 3:}]\ \ Output the final result of information volume of mass function $E_i(\varepsilon)$.
\end{description}

The Information Volume of Mass Function (IVMF) in D-S evidence theory quantifies the granularity and uncertainty embedded within evidence, serving as a critical metric to evaluate the informativeness of belief assignments across hypotheses. In fields such as complex system diagnostics and multi-sensor networks, IVMF optimizes evidence fusion by prioritizing high-information mass functions (e.g., those with lower entropy or sharper belief distributions), thereby enhancing decision reliability in scenarios like fault detection or target tracking. In environmental monitoring, IVMF guides adaptive data collection by identifying regions or parameters where uncertainty dominates, enabling efficient resource allocation. For semi-supervised medical applications, IVMF plays a transformative role by dynamically weighting unlabeled samples based on their evidential richness. For instance, in medical image segmentation, IVMF evaluates the ambiguity of pixel-level predictions—assigning higher weights to regions where mass functions exhibit concentrated belief (e.g., clear tumor boundaries) while deprioritizing areas with diffuse or conflicting evidence (e.g., ambiguous lesions). This approach reduces noise in pseudo-labels and strengthens consistency regularization, particularly in low-data regimes. Moreover, IVMF-driven learning frameworks selectively query expert annotations for samples with maximal informational value (e.g., mass functions straddling decision boundaries), accelerating model convergence and improving diagnostic robustness. By integrating evidential uncertainty quantification with semi-supervised paradigms, IVMF serves as a reliable metric that enables medical AI systems to more effectively utilize sparse labeled data, while preserving interpretability in critical tasks such as early disease detection or rare pathology classification.

\subsection{Training Strategy of the Two Sub-Networks}
For the main architecture of networks, it consists of two sub-networks in which both of them $\mathcal{N}_{s1}$ and $\mathcal{N}_{s2}$ teach each other to learn. For the pre-training stage, the sub-networks are trained using labeled data to obtain corresponding model parameters $\mathcal{I}_{s1}$ and $\mathcal{I}_{s2}$. In the self-training stage, the sub-network $\mathcal{N}_{s2}$ is supposed to update its parameters after the sub-network $\mathcal{N}_{s1}$ carries out the back-propagating operation. Moreover, the training strategy is the same as BCP \cite{DBLP:conf/cvpr/BaiCL0023}.

\begin{figure*}
	\centering 
	\includegraphics[width = 17.5cm]{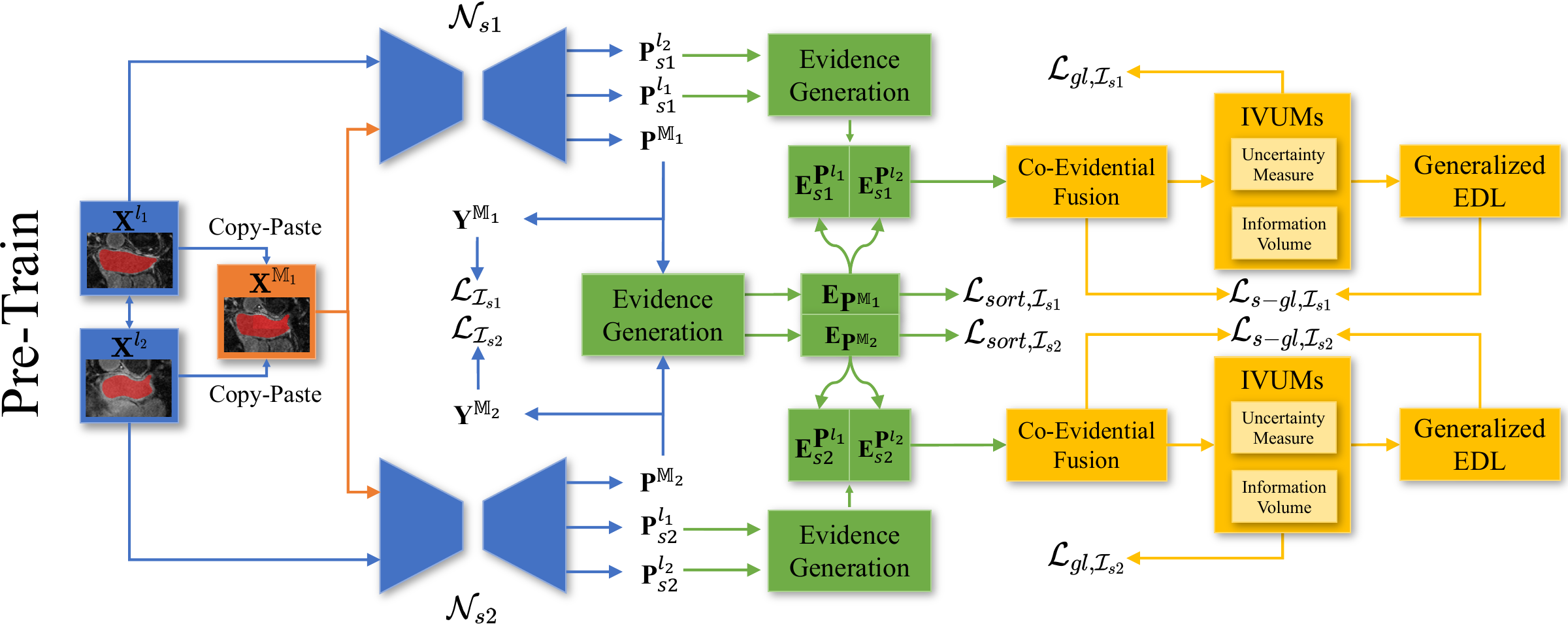}
	\caption{Details of the Pre-training Stage. \underline{\textit{L–LM}} : Labeled and Labeled Mixed Information.}
	\label{pre}
\end{figure*}

\subsection{Evidential Pre-training Stage}
The details of the pre-training stage are provided in Fig. \ref{pre}. In the stage of pre-training, we initiate two random zero-centered masks $\mathcal{M}_1, \mathcal{M}_2 \in \{0,1\}^{W\times H \times L}$ which are utilized to represent whether a voxel is located in the foreground (0) or the background (1) image. To be specific, the size of zero-value region is $\eta W\times\eta H \times\eta L, \eta \in (0,1)$ and the process of copy-pasting for obtaining model parameter $\mathcal{I}_{s1}$ with mask $\mathcal{M}_1$ can be defined as:
\begin{equation}
	\textbf{X}^{\mathbb{M}_{1}} = \textbf{X}^{l_1}_a \odot \mathcal{M}_1 + \textbf{X}^{l_2}_b \odot (\mathbf{1} - \mathcal{M}_1)
\end{equation}

\begin{equation}
	\textbf{Y}^{\mathbb{M}_{1}} = \textbf{Y}^{l_1}_a \odot \mathcal{M}_1 + \textbf{Y}^{l_2}_b \odot (\mathbf{1} - \mathcal{M}_1)
\end{equation}
where $\mathbf{1} \in \{1\}^{W\times H \times L}$, $a\neq b$ and $\odot$ represents elemnent-wise multiplication. Similarly, another model parameter $\mathcal{I}_{s2}$ can be obtained utilizing $\textbf{X}^{\mathbb{M}_{2}}$ and $\textbf{Y}^{\mathbb{M}_{2}}$ with the same copy-paste strategy utilizing mask $\mathcal{M}_2$. The predictions generated by the models can be given as:
\begin{equation}
	\textbf{P}^{\mathbb{M}_{1}} = \mathcal{N}(\textbf{X}^{\mathbb{M}_{1}}), \quad \textbf{P}^{\mathbb{M}_{2}} = \mathcal{N}(\textbf{X}^{\mathbb{M}_{2}})
\end{equation}

Therefore, we are able to construct the base loss function of pre-training stage for parameters $\mathcal{I}_{s1}$ and $\mathcal{I}_{s2}$ which can be defined as:
\begin{equation}
	\mathcal{L}_{\mathcal{I}_{s1}} = \mathcal{L}_{ce}(\textbf{P}^{\mathbb{M}_{1}},\textbf{Y}^{\mathbb{M}_{1}}) + \mathcal{L}_{dice}(\textbf{P}^{\mathbb{M}_{1}},\textbf{Y}^{\mathbb{M}_{1}})
\end{equation}
\begin{equation}
	\mathcal{L}_{\mathcal{I}_{s2}} = \mathcal{L}_{ce}(\textbf{P}^{\mathbb{M}_{2}},\textbf{Y}^{\mathbb{M}_{2}}) + \mathcal{L}_{dice}(\textbf{P}^{\mathbb{M}_{2}},\textbf{Y}^{\mathbb{M}_{2}})
\end{equation}
where $\mathcal{L}_{ce}$ and $\mathcal{L}_{dice}$ represent Cross Entropy and DICE loss, respectively. And the predictions $\textbf{P}^{\mathbb{M}_{1}}$ and $\textbf{P}^{\mathbb{M}_{2}}$ are expected to be transformed into evidences referencing evidence generation procedure proposed in previous work \cite{DBLP:conf/cvpr/ChenGX23}. The transformation can be defined as:
\begin{equation}
	\textbf{E}_{\textbf{P}^{\mathbb{M}_{1}}} = \textbf{G}(\textbf{P}^{\mathbb{M}_{1}}), \quad \textbf{E}_{\textbf{P}^{\mathbb{M}_{2}}} = \textbf{G}(\textbf{P}^{\mathbb{M}_{2}})
\end{equation}
where $\textbf{E}_{\textbf{P}^{\mathbb{M}_{1}}}$ and $\textbf{E}_{\textbf{P}^{\mathbb{M}_{2}}}$ denote transformed evidences and $\textbf{G}$ represents the evidence generation procedure, $e.g.$ ReLu, which ensures the elements in the generated evidence non-negative. The dimension of class is selected as the position where the evidences are generated, which indicates that the total number of evidences for a sample equals to $W\times H \times L$. And a piece of evidence corresponding to a voxel reflects the confidence of model on its prediction.

\subsubsection{Uncertainty Guided Learning (UGL)} In order to fully illustrate the usage of evidences, an evidence vector is abstracted among the ones for parameter $\mathcal{I}_{s1}$ generated in a sample $\psi_f, f \in [1, \mathbb{F}]$:
\begin{equation}
	\mathbf{v}_{\eta} = [v_{\eta,1},...,v_{\eta,N}]
\end{equation}
where $\mathbb{H}$ is the total number of samples, $N$ represents the dimension size of classes and $\eta \in [1, W\times H \times L]$. For each evidence generated with respect to a sample, it is supposed to be calculated to obtain a corresponding uncertainty measure, which can be given as:
\begin{equation}
	\mathcal{U}_{\psi_f} = [ \mathcal{U}_{\psi_f,1}, ..., \mathcal{U}_{\psi_f,\eta}, ..., \mathcal{U}_{\psi_f,W\times H \times L} ]
\end{equation}
where $\mathcal{U}_{\psi_f}$ denotes complete uncertainty measure for every voxel in sample $\psi_f$. When all the uncertainty measures of samples are obtained, they can be denoted by $\Upsilon$. For each generated piece of evidence, it is able to express the level of difficulty in learning features of a voxel. To better guide model to focus on hard parts of features, we propose to attach higher weights to voxels with a lower degree of uncertainty at first and then increase weights of voxels possessing a higher degree of uncertainty. In this way, the model is able to focus on easy-to-learn voxels and pay more attention to difficult ones. To be specific, the dynamic weight allocation function can be defined as:
\begin{equation}
	\varpi(r,t) = \phi \cdot tanh(\xi(r)\lambda(h(t))) + 1
\end{equation}
where $\phi$ is designed to restrict the amplitude of change of weight and $\xi(r) = \frac{2r}{R} - 1 \in [-1,1], r=1,...,R$, $R$ represents the total number of training epochs and $r$ denotes the index of current epoch. Besides, $\lambda(h(t)) = \frac{2h(t)}{T} - 1 \in [-1,1], t = 1,...,T$, $h(t)$ represents the ordinal number of each voxel which is acquired by sorting $\Upsilon$ in a descending order. For the stage of pre-training, the uncertainty sort loss function for obtaining parameter $\mathcal{I}_s$ can be given as:
\begin{equation}
	\mathcal{L}_{sort,\mathcal{I}_{s1}} = \sum_{f=1}^{\mathbb{F}}\sum_{\eta=1}^{W\times H \times L}\frac{\varpi(r,t)}{W\times H \times L}\cdot\mathcal{L}_{cd}(\textbf{P}^{^{\mathbb{M}_{1}}}_{\psi_h^{\mathcal{U}_{\eta}}},\textbf{Y}^{^{\mathbb{M}_{1}}}_{\psi_h^{\mathcal{U}_{\eta}}})
\end{equation}
where $\mathcal{L}_{cd}$ represents the combination of $\mathcal{L}_{ce}$ and $\mathcal{L}_{dice}$. Besides, $\textbf{P}^{^{\mathbb{M}_{1}}}_{\psi_h^{\mathcal{U}_{\eta}}}$  and $\textbf{Y}^{^{\mathbb{M}_{1}}}_{\psi_h^{\mathcal{U}_{\eta}}}$ represent the prediction vector and ground truth which are located in the same position of evidence where $\mathcal{U}_{\eta}$ is generated. The parameter $\mathcal{I}_{s2}$ can be obtained in the same way using the pre-processed data using mask $\mathcal{M}_2$. 
\subsubsection{Co-Evidential Fusion Design with Information Volume of Mass Function (CDIF)}
Based on the definition of the traditional optimization target, the optimization process is mainly improved by utilizing information volume of mass function. For each generated evidence with corresponding uncertainty measure, $\mathbf{v}_{\eta}$ is taken for an example:
\begin{equation}
	\mathbf{v}_{\eta}^{\mathcal{U}} = [v_{\eta,1},...,v_{\eta,N}, \mathcal{U}_{\psi_f,\eta}]
\end{equation}
where $\mathbf{v}_{\eta}^{\mathcal{U}}$ is the complete form of the generated evidence. For the traditional framework of D-S evidence theory, $\mathbf{v}_{\eta}$ can be rewritten as:
\begin{equation}
	v_{\eta,i} = \varepsilon_1(P_i^{\eta}),\quad \mathcal{U}_{\psi_f,\eta} = \varepsilon_1(P_{N+1}^{\eta})
\end{equation}
where $i \in [1,N]$. In order to enable the model to learn further by referencing guiding information contained in labeled data, the original labeled data is input into model to generate other groups of evidences. For obtaining parameter $\mathcal{I}_{s1}$, the process of generating corresponding predictions can be defined as:
\begin{equation}
	\textbf{P}^{{l_1}}_{s1} = \mathcal{N}_{s1}(\textbf{X}^{{l_1}}), \quad \textbf{P}^{{l_2}}_{s1} = \mathcal{N}_{s1}(\textbf{X}^{{l_2}})
\end{equation}

In the same way, the predictions are produced by sub-network $\mathcal{N}_{s2}$ using mask $\mathcal{M}_2$. Then, utilizing the evidence transformation process, $\textbf{P}^{{l_1}}_{s1}$ and $\textbf{P}^{{l_2}}_{s1}$ can be transformed into evidences whose process can be given as:
\begin{equation}
	\textbf{E}^{\textbf{P}^{{l_1}}}_{s1} = \textbf{G}(\textbf{P}^{{l_1}}), \quad \textbf{E}^{\textbf{P}^{{l_2}}}_{s1} = \textbf{G}(\textbf{P}^{{l_2}})
\end{equation}

$\textbf{E}^{\textbf{P}^{{l_1}}}_{s2}$ and $\textbf{E}^{\textbf{P}^{{l_1}}}_{s2}$ can be also obtained. Similarly, we abstract the evidences contained $\textbf{E}^{\textbf{P}^{{l_1}}}_{s1}$ and $\textbf{E}^{\textbf{P}^{{l_2}}}_{s1}$ which possess the same position of $\mathbf{v}_{\eta}^{\mathcal{U}}$:
\begin{equation}
	\begin{aligned}
		&\mathbf{v}_{\eta'}^{\mathcal{U}'} = [v_{\eta',1},...,v_{\eta',N}, \mathcal{U}_{\psi_f,\eta'}],\\
		&\mathbf{v}_{\eta''}^{\mathcal{U}''} = [v_{\eta'',1},...,v_{\eta'',N}, \mathcal{U}_{\psi_f,\eta''}]
	\end{aligned}
\end{equation}

The corresponding form of evidences in the traditional framework of D-S evidence theory can be given as:
\begin{equation}
	\begin{aligned}
		&v_{\eta',i} = \varepsilon_2(P_i^{\eta}),& \mathcal{U}_{\psi_f,\eta'} = \varepsilon_2(P_{N+1}^{\eta})\\
		&v_{\eta'',i} = \varepsilon_3(P_i^{\eta}),& \mathcal{U}_{\psi_f,\eta''} = \varepsilon_3(P_{N+1}^{\eta})
	\end{aligned}
\end{equation}

Here, for a more efficient fusion of evidences and generation of a more accurate uncertainty measure, we propose a novel evidence combination rule based on the pignistic transformation which can be defined as:
\begin{equation}
	\begin{aligned}
		\begin{split}
			&\varepsilon_{\varsigma}(\{P_i\}) = (\varepsilon_d(\{P_i\}) * \varepsilon_e(\{P_i\}) +\frac{|P_i|}{|\mathcal{S}|+|P_i|}*\\  &(\varepsilon_d(\{P_i\}) * \varepsilon_e(\{\mathcal{S}\}) + \varepsilon_d(\{\mathcal{S}\})* \varepsilon_e(\{P_i\}))&
		\end{split}
	\end{aligned}
\end{equation}
\begin{equation}
	\varepsilon_{final}(\{P_i\}) = \frac{\varepsilon_{\varsigma}(\{P_i\})}{\sum_{i = 1}^{N+1}\varepsilon_{\varsigma}\{P_i\}}
\end{equation}
where $\varepsilon_d$ and $\varepsilon_e$ represent two piece of evidences. In the process of evidence fusion, the masses of different propositions will be affected by varying degrees. For the propositions with larger cardinalities, their belief values will be subject to much smaller changes which remain more similar to their original conditions, which is designed to obtain more obvious uncertainty indicators. For simplifying the process of evidence fusion, $\mathcal{F}$ is utilized to denote the fusion procedure. The predictions from two groups of original labeled data are expected to be fused with the ones from mixed data respectively:
\begin{equation}
	\varepsilon_{1,2} = \mathcal{F}(\varepsilon_{1},\varepsilon_{2}), \quad \varepsilon_{1,3} = \mathcal{F}(\varepsilon_{1},\varepsilon_{3})
\end{equation}
where $\varepsilon_{1,2}$ and $\varepsilon_{1,3}$ represent fused results of $\varepsilon_{1}$, $\varepsilon_{2}$ and $\varepsilon_{2}$, $\varepsilon_{3}$. In the view of subjective logic-based evidence theory, each evidence possesses a uncertainty measure naturally which can be utilized as a reference of the understanding level of the model for each voxel feature. However, in the framework of traditional D-S evidence theory, the uncertainty measure includes evaluation of multiple sub-sets, namely the original uncertainty representation in subjective logic-based evidence theory. We propose a novel evidence evaluation method called $\textbf{IVUM}$ synthesizing the concept of information volume of mass function and original uncertainty measure. For $\varepsilon_{1,2}$, the corresponding $\textbf{IVUM}$ can be given as:
\begin{equation}
	\textbf{IVUM}_{\varepsilon_{1,2}} = \varepsilon_{1,2}(P_{N+1}^{\eta})*\mathcal{I}\mathcal{V}(\varepsilon_{1,2})
\end{equation}
where $\mathcal{I}\mathcal{V}$ represents the process of generating information volume of $\varepsilon_{1,2}$. Similarly, $\textbf{IVUM}_{\varepsilon_{2,3}}$ can be obtained in the same way. Then, we obtain fused evidences $\textbf{E}_{\mathcal{F}_1}$ and $\textbf{E}_{\mathcal{F}_2}$ using $\textbf{E}_{\textbf{P}^{\mathbb{M}_{1}}}$ which is expected to be combined with $\textbf{E}_{\textbf{P}^{{l_1}}}$ and $\textbf{E}_{\textbf{P}^{{l_2}}}$. For the fused evidences, it is supposed to calculate the two groups of $\textbf{IVUM}$s of corresponding evidences to serve as references to guide model to learn voxel-wisely which are represented by $\textbf{IVUM}_1$ and $\textbf{IVUM}_2$ respectively. To simplify the computational difficulty, we normalize $\textbf{IVUM}_1$ and $\textbf{IVUM}_2$, which is denoted by $\widetilde{\textbf{IVUM}_1}$ and $\widetilde{\textbf{IVUM}_2}$. With respect the traditional evidential deep learning, we improve the corresponding optimization target under the generalized framework of EDL using $\widetilde{\textbf{IVUM}}$:
\begin{equation}
	\mathcal{L}_{gl,\mathcal{I}_{s1}} = \sum_{j=1}^{M}(1-\widetilde{\textbf{IVUM}}^j)\sum_{k=1}^{N}y_k^j(log \mathcal{U}^j-log c_k^j)
\end{equation}
where $\mathcal{L}_{gl}$ represents the improved evidential learning of two groups of fused evidences $\textbf{E}_{\mathcal{F}_1}$ and $\textbf{E}_{\mathcal{F}_2}$. 

\subsubsection{Uncertainty Guided Learning with IVUM (ULM)} Then, by sorting $\widetilde{\textbf{IVUM}}$, a new order can be obtained which is different from the sequence acquired by sorting $\Upsilon$. The newly dynamic weight allocation function can be defined as:
\begin{equation}
	\varpi(r,t)_{\widetilde{\textbf{IVUM}}} = \phi \cdot tanh(\xi(r)\lambda(h(t_{\widetilde{\textbf{IVUM}}}))) + 1
\end{equation}

By synthesizing the newly dynamic weight allocation function and the improved optimization target, a more detailed fine-grained learning pattern can be designed:
\begin{equation}
	\mathcal{L}_{s-gl,\mathcal{I}_{s1}} = \varpi(r,t)_{\widetilde{\textbf{IVUM}}}*\mathcal{L}_{gl}
\end{equation}

Therefore, for obtaining the parameter $\mathcal{I}_{s1}$, the final total loss function can be defined as:
\begin{equation}
	\mathcal{L}_{\mathcal{I}_{s1}}^{final} = \mathcal{L}_{\mathcal{I}_s1}+\lambda_1\mathcal{L}_{sort,\mathcal{I}_{s1}}+\lambda_2\mathcal{L}_{gl,\mathcal{I}_{s1}}+\lambda_3\mathcal{L}_{s-gl,\mathcal{I}_{s1}}
\end{equation}
where $\lambda_1$, $\lambda_2$ and $\lambda_3$ are balancing parameters. The evidential learning process that produces $\lambda_1\mathcal{L}_{sort,\mathcal{I}}+\lambda_2\mathcal{L}_{gedl,\mathcal{I}}+\lambda_3\mathcal{L}_{s-gl,\mathcal{I}}$ is denoted by $\mathbb{E}$. For the acquisition of parameter $\mathcal{I}_{s2}$, the corresponding optimization objective can be defined as:
\begin{equation}
	\mathcal{L}_{\mathcal{I}_{s2}}^{final} = \mathcal{L}_{\mathcal{I}_{s1}} + \mathbb{E}_{\mathcal{I}_{s1}}
\end{equation}
where $\mathcal{L}_{\mathcal{I}_{s2}}^{final}$ represents the final objective function for $\mathcal{N}_{s2}$ during the pre-training stage. Additionally, it is worth noting that $\mathcal{L}_{\mathcal{I}_{s1}}^{final}$ is the final loss function for $\mathcal{N}_{s1}$ during the pre-training stage.
\begin{figure*}
	\centering 
	\includegraphics[width = 17.5cm]{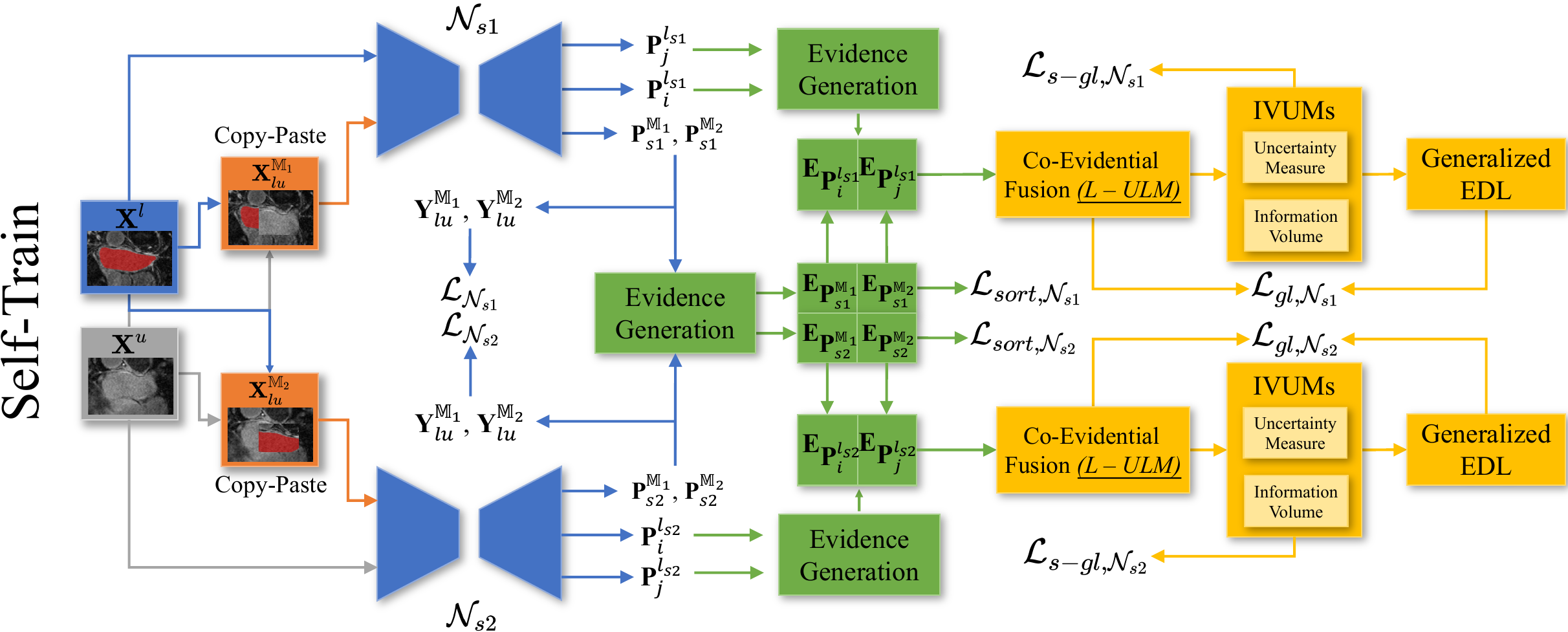}
	\caption{Details of the Self-training Stage. \underline{\textit{L–ULM}} :Labeled and Unlabeled \& Labeled Mixed Information.}
	\label{self}
\end{figure*}
\subsection{Evidential Self-training Stage}
The details of the self-training stage are provided in Fig. \ref{self}. Similar to the data pre-processing procedure of pre-training stage, the operation is carried out between labeled and unlabeled data, which aims to help model to learn semantic relations between them. Specifically, the input which is supposed to be processed by mask $\mathcal{M}$ can be defined as:
\begin{equation}
	\textbf{X}_{lu}^{\mathbb{M}_1} = \textbf{X}_{i}^{l} \odot \mathcal{M} + \textbf{X}_{m}^{u} \odot (\textbf{1}-\mathcal{M})
\end{equation}
\begin{equation}
	\textbf{X}_{lu}^{\mathbb{M}_2} = \textbf{X}_{j}^{l} \odot (\textbf{1}-\mathcal{M}) + \textbf{X}_{n}^{u} \odot \mathcal{M}
\end{equation}

Obviously, the unlabeled data dose not possess supervised signals. To generate the supervised signals, unlabeled part of data is fed into sub-network and then the predictions are utilized to train another network. For example, take the sub-network $\mathcal{N}_{s1}$ as teacher, the process can be given as:
\begin{equation}
	\textbf{P}_{m}^{u} = \mathcal{N}_{s2}(\textbf{X}_{m}^{u}, \mathcal{I}_{s2}), \quad\textbf{P}_{n}^{u} = \mathcal{N}_{s2}(\textbf{X}_{n}^{u}, \mathcal{I}_{s2})
\end{equation}

$\textbf{P}^{u}$ is utilized to generate initial pseudo-label (IPL) and the one $\textbf{Y}^{u}$ which is used in the end are produced by choosing the largest connected component of IPL. Then, for the supervisory signal, the same operation is carried out to produce corresponding labels, which can be defined as:
\begin{equation}
	\textbf{Y}_{lu}^{\mathbb{M}_1} = \textbf{Y}_{i}^{l}\odot \mathcal{M}+\textbf{Y}^{u}_m \odot (\textbf{1}-\mathcal{M})
\end{equation}
\begin{equation}
	\textbf{Y}_{lu}^{\mathbb{M}_2} = \textbf{Y}_{j}^{l}\odot (\textbf{1}-\mathcal{M})+\textbf{Y}^{u}_n\odot \mathcal{M}
\end{equation}
\begin{figure*}[htbp]
	\centering
	\begin{minipage}{0.49\linewidth}
		\centering
		\includegraphics[width=9cm]{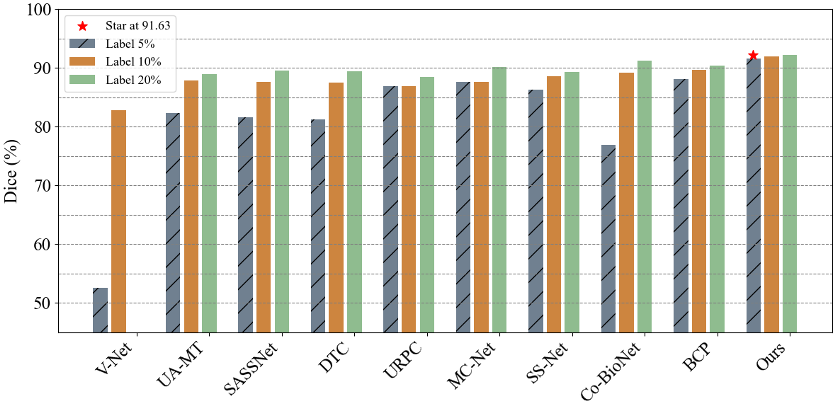}
		\caption{The Dice Results of the Comparative Models on LA Dataset in $5\%$, $10\%$ and $20\%$ Labeled Ratios.}
		\label{q}
	\end{minipage}
	\begin{minipage}{0.49\linewidth}
		\centering
		\includegraphics[width=9cm]{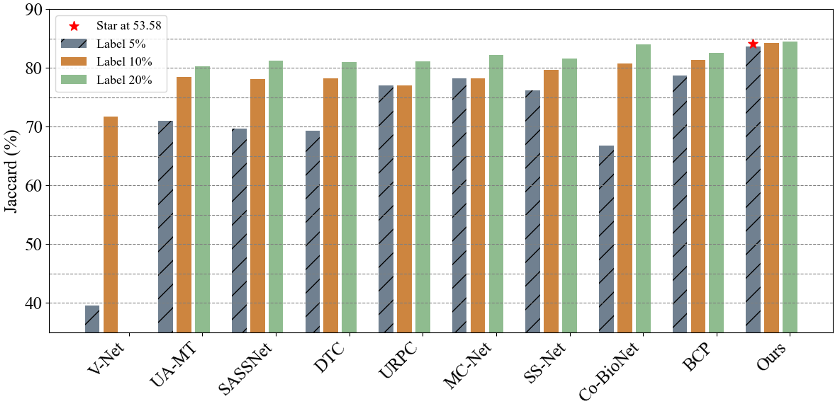}
		\caption{The Jaccard Results of the Comparative Models on LA Dataset in $5\%$, $10\%$ and $20\%$ Labeled Ratios.}
		\label{w}
	\end{minipage}
	\vspace{0.2cm}
	
	\begin{minipage}{0.49\linewidth}
		\centering
		\includegraphics[width=9cm]{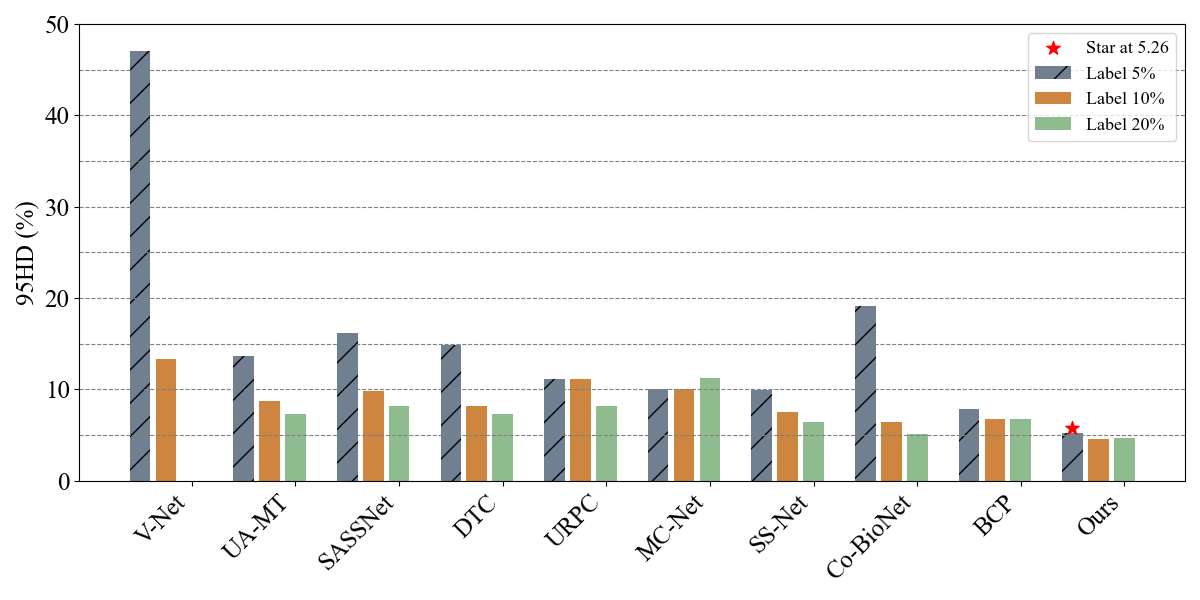}
		\caption{The 95HD Results of the Comparative Models on LA Dataset in $5\%$, $10\%$ and $20\%$ Labeled Ratios.}
		\label{e}
	\end{minipage}
	\begin{minipage}{0.49\linewidth}
		\centering
		\includegraphics[width=9cm]{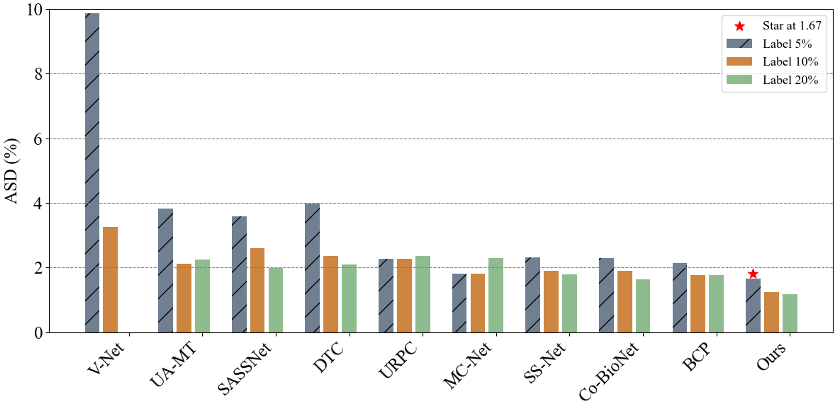}
		\caption{The ASD Results of the Comparative Models on LA Dataset in $5\%$, $10\%$ and $20\%$ Labeled Ratios.}
		\label{r}
	\end{minipage}
\end{figure*}
With the mixed input $\textbf{X}_{lu}^{\mathbb{M}_1}$ and $\textbf{X}_{lu}^{\mathbb{M}_2}$, both of them can be fed into the two sub-networks to generate corresponding predictions to serve for network optimizations. To be specific, for the sub-network $\mathcal{N}_{s1}$, the predictions can be given as:
\begin{equation}
	\textbf{P}_{s1}^{\mathbb{M}_1} = \mathcal{N}_{s1}(\textbf{X}_{lu}^{\mathbb{M}_1} , \mathcal{I}_{s1}), \quad\textbf{P}_{s1}^{\mathbb{M}_2} = \mathcal{N}_{s1}(\textbf{X}_{lu}^{\mathbb{M}_2}, \mathcal{I}_{s1})
\end{equation}

For the sub-network $\mathcal{N}_{s2}$, the predictions are $\textbf{P}_{s2}^{\mathbb{M}_1}$ and $\textbf{P}_{s2}^{\mathbb{M}_2}$. Next, we are able to construct base loss function of self-training stage for the optimization of two sub-networks $\mathcal{N}_{s1}$ and $\mathcal{N}_{s2}$ which can be defined as:
\begin{equation}
	\begin{aligned}
		&\mathcal{L}_{\mathcal{N}_{s1}} &= \mathcal{L}_{ce}(\textbf{P}_{s1}^{\mathbb{M}_1},\textbf{Y}_{lu}^{\mathbb{M}_1}) + \mathcal{L}_{dice}(\textbf{P}_{s1}^{\mathbb{M}_1},\textbf{Y}_{lu}^{\mathbb{M}_1})\\ && + \mathcal{L}_{ce}(\textbf{P}_{s1}^{\mathbb{M}_2},\textbf{Y}_{lu}^{\mathbb{M}_2}) + \mathcal{L}_{dice}(\textbf{P}_{s1}^{\mathbb{M}_2},\textbf{Y}_{lu}^{\mathbb{M}_2})
	\end{aligned}
\end{equation}

For the sub-network $\mathcal{N}_{s2}$, the optimization $\mathcal{L}_{s2}$ can be obtained in the same way.

\subsubsection{UGL in Self-training Stage} For the self-training stage, we design the fusion strategy that the original predictions from mixed input are expected to reference information from labeled data, which further assists the model in learning semantic relations more effectively. First, the predictions of mixed input are transformed into evidences which can be given as:
\begin{equation}
	\textbf{E}_{\textbf{P}_{s1}^{\mathbb{M}_1}} = \textbf{G}(\textbf{P}_{s1}^{\mathbb{M}_1}), \quad \textbf{E}_{\textbf{P}_{s1}^{\mathbb{M}_2}} = \textbf{G}(\textbf{P}_{s1}^{\mathbb{M}_2})
\end{equation}

With respect to sub-network $\mathcal{N}_{s2}$, $\textbf{E}_{\textbf{P}_{s2}^{\mathbb{M}_1}}$ and $\textbf{E}_{\textbf{P}_{s2}^{\mathbb{M}_2}}$ can also be calculated. For the evidences obtained above, each of them is supposed to possess a degree of uncertainty. The process of obtaining uncertainty sort loss is denoted by $\mathcal{R}\mathcal{T}$ and the ones for self-training stage can be given as:
\begin{equation}
	\mathcal{L}_{sort,\mathcal{N}_{s1}} = \mathcal{R}\mathcal{T}(\textbf{E}_{\textbf{P}_{s1}^{\mathbb{M}_1}})+ \mathcal{R}\mathcal{T}(\textbf{E}_{\textbf{P}_{s1}^{\mathbb{M}_2}})
\end{equation}

For the uncertainty sort loss of sub-network $\mathcal{N}_{s2}$, $\mathcal{L}_{sort,\mathcal{N}_{s2}}$ can also be acquired.

\subsubsection{CDIF in Self-training Stage} Then, for further utilization of labeled information to assist model in learning the underlying semantic relation between labeled and unlabeled data, the labeled data is supposed to be fed into the networks to generate predictions first, which can be given as:
\begin{equation}
	\textbf{P}_{i}^{l_{s1}} = \mathcal{N}_{s1}(\textbf{X}_{i}^{l}, \mathcal{I}_{s1}), \quad\textbf{P}_{j}^{l_{s1}} = \mathcal{N}_{s1}(\textbf{X}_{j}^{l}, \mathcal{I}_{s1})
\end{equation}

$\textbf{P}_{i}^{l_{s2}}$ and $\textbf{P}_{j}^{l_{s2}}$ can be computed. The predictions $\textbf{P}_{ij}^{l_{s1}}$ and $\textbf{P}_{ij}^{l_{s2}}$ are transformed into evidences utilizing the aforentioned evidences generation procedure, which can be given as:
\begin{equation}
	\textbf{E}_{\textbf{P}_{i}^{l_{s1}}} = \textbf{G}(\textbf{P}_{i}^{l_{s1}}), \quad \textbf{E}_{\textbf{P}_{j}^{l_{s1}}} = \textbf{G}(\textbf{P}_{j}^{l_{s1}})
\end{equation}

For the optimization of two sub-networks, the process of obtaining fused evidences can be given as:
\begin{equation}
	\textbf{E}_{s1}^{\mathcal{F}_{1}} = \mathcal{F}(\textbf{E}_{\textbf{P}_{lu}^{\mathbb{M}_1}}, \textbf{E}_{\textbf{P}_{i}^{l_{s1}}}), \quad \textbf{E}_{s1}^{\mathcal{F}_{2}} = \mathcal{F}(\textbf{E}_{\textbf{P}_{lu}^{\mathbb{M}_2}}, \textbf{E}_{\textbf{P}_{j}^{l_{s2}}})
\end{equation}

Evidences $\textbf{E}_{s2}^{\mathcal{F}_{1}}$ and $\textbf{E}_{s2}^{\mathcal{F}_{2}}$ for sub-network $\mathcal{N}_{s2}$ can be obtained in the same way. With the fused evidences, the process of constructing the improved evidential optimization loss with generated $\textbf{IVUM}$s is denoted as $\mathcal{I}\mathcal{G}$ and corresponding loss functions for the sub-networks can be given as:
\begin{equation}
	\mathcal{L}_{gl,\mathcal{N}_{s1}} = \mathcal{I}\mathcal{G}(\textbf{E}_{s1}^{\mathcal{F}_{1}}) + \mathcal{I}\mathcal{G}(\textbf{E}_{s1}^{\mathcal{F}_{2}})
\end{equation}
Besides, $\mathcal{L}_{gl,\mathcal{N}_{s2}}$ can also be calculated.

\subsubsection{ULM in Self-training Stage} On the basis of obtained $\textbf{IVUM}$s, the process of producing $\mathcal{L}_{s-gl}$ is denoted by $\mathcal{G}\mathcal{D}$ and the $\mathcal{L}_{s-gl}$s for self-training stage can be given as:
\begin{equation}
	\mathcal{L}_{s-gl,\mathcal{N}_{s1}} = \mathcal{G}\mathcal{D}(\textbf{E}_{s1}^{\mathcal{F}_{1}}) + \mathcal{G}\mathcal{D}(\textbf{E}_{s1}^{\mathcal{F}_{2}})
\end{equation}

$\mathcal{L}_{s-gl,\mathcal{N}_{s2}}$ for sub-network $\mathcal{N}_{s2}$ can be obtained accordingly. In the end, take sub-network $\mathcal{N}_{s1}$ as an example, and the total loss function for the sub-network can be given as:
\begin{equation}
	\mathcal{L}_{\mathcal{N}_{s1}}^{final} = \mathcal{L}_{\mathcal{N}_{s1}}+\lambda_4\mathcal{L}_{sort,\mathcal{N}_{s1}}+\lambda_5\mathcal{L}_{gl,\mathcal{N}_{s1}}+\lambda_6\mathcal{L}_{s-gl,\mathcal{N}_{s1}}
\end{equation}

For another sub-network, the corresponding optimization objective can be defined as:
\begin{equation}
	\mathcal{L}_{\mathcal{N}_{s2}}^{final} = \mathcal{L}_{\mathcal{N}_{s2}}+\lambda_4\mathcal{L}_{sort,\mathcal{N}_{s2}}+\lambda_5\mathcal{L}_{gl,\mathcal{N}_{s2}}+\lambda_6\mathcal{L}_{s-gl,\mathcal{N}_{s2}}
\end{equation}
where $\mathcal{L}_{\mathcal{N}_{s1}}^{final}$ and $\mathcal{L}_{\mathcal{N}_{s2}}^{final}$ share the same set of balancing parameters $\lambda_4, \lambda_5, \lambda_6$, and represent the final objective function for sub-networks $\mathcal{N}_{s1}$ and $\mathcal{N}_{s2}$ during the self-training stage. It is worth noting that sub-network $\mathcal{N}_{s1}$ updates its parameters prior to the optimization of sub-network $\mathcal{N}_{s2}$, meaning that sub-network $\mathcal{N}_{s1}$ generates pseudo-labels after its backpropagation process.

\begin{table*}[htbp] \scriptsize
	\renewcommand{\arraystretch}{1.2}
	\label{table5}
	\centering
	\caption{Comparison with state-of-the-art models on LA, NIH-Pancreas and ACDC datasets in labeled ratio 5$\%$, 10$\%$ and 20$\%$}
	\begin{threeparttable}
		\setlength{\tabcolsep}{1.1mm}{
			\begin{tabular}{|c |c c |c c c c| c c c c| c c c c|}
				\hline
				\multicolumn{1}{|c|}{\multirow{2}*{Dataset}}& \multicolumn{2}{c|}{\multirow{2}*{Scans Used}} &\multicolumn{4}{c|}{LA Dataset}& \multicolumn{4}{c|}{NIH-Pancreas Dataset}& \multicolumn{4}{c|}{ACDC Dataset}\\\cline{4-15}
				{}&{} &{} &\multicolumn{4}{c|}{Metrics}&\multicolumn{4}{c|}{Metrics}&\multicolumn{4}{c|}{Metrics}  \\\cline{1-15}
				\multicolumn{1}{|c|}{\multirow{1}*{Model}}&{Labeled} & {Unlabled} & Dice$\uparrow$ & Jaccard$\uparrow$ & 95HD$\downarrow$ & ASD$\downarrow$& Dice$\uparrow$ & Jaccard$\uparrow$ & 95HD$\downarrow$ & ASD$\downarrow$& Dice$\uparrow$ & Jaccard$\uparrow$ & 95HD$\downarrow$ & ASD$\downarrow$\\ \cline{1-15}
				V-Net / U-net$^{*}$&(5\%)&(95\%)& 52.55 & 39.60 & 47.05 & 9.87&55.06&40.48&32.8&12.67& 47.83 & 37.01 & 31.16 & 12.62\\
				V-Net / U-net$^{*}$&(10\%)&(90\%)& 82.74 & 71.72 & 13.35 & 3.26&69.65&55.19&20.2&6.31& 79.41 & 68.11 & 9.35 & 2.70\\
				V-Net / U-net$^{*}$&(100\%)&(0\%)& \textbf{92.62} & \textbf{85.24} & \textbf{4.47} & \textbf{1.33}&\textbf{85.74}&\textbf{73.86}&\textbf{4.48}&\textbf{1.07}& \textbf{92.57} & \textbf{85.31} & \textbf{2.91} & \textbf{0.74}\\\hline
				UA-MT&\multirow{9}*{(5\%)} &\multirow{9}*{(95\%)}& 82.26 & 70.98 & 13.71 & 3.82&47.03&32.79&35.31&4.26& 46.04 & 35.97 & 20.08 & 7.75 \\
				SASSNet&&& 81.60 & 69.63 & 16.16 & 3.58&56.05&41.56&36.61&4.90& 57.77 & 46.14 & 20.05 & 6.06 \\
				DTC&&& 81.25 & 69.33 & 14.90 & 3.99&49.83&34.47&41.16&16.53& 56.90 & 45.67 & 23.36 & 7.39 \\
				URPC&&&86.92 & 77.03 & 11.13 & 2.28&52.05&36.47&34.02&13.16& 55.87 & 44.64 & 13.60 & 3.74 \\
				MC-Net&&& 87.62 & 78.25 & 10.03 & 1.82&54.99&40.65&16.03&3.87& 62.85 & 52.29 & 7.62 & 2.33\\
				SS-Net&&&86.33 & 76.15 & 9.97 & 2.31&56.35&43.41&22.75&5.39&65.83 & 55.38 & 6.67 & 2.28\\
				Co-BioNet&&&76.88&66.76&19.09&2.30&79.74&65.66&5.43&2.79&88.05&79.61&4.22&1.18\\
				BCP&&& 88.02& 78.72 & 7.90 & 2.15&80.33&67.65&11.78&4.32& 87.59& 78.67& \textcolor{red}{\textbf{1.90}}& \textcolor{red}{\textbf{0.67}} \\
				Ours&&&\textcolor{red}{\textbf{91.63}}& \textcolor{red}{\textbf{83.58}} & \textcolor{red}{\textbf{5.26}} & \textcolor{red}{\textbf{1.67}}&\textcolor{red}{\textbf{82.36}}&\textcolor{red}{\textbf{70.38}}&\textcolor{red}{\textbf{7.67}}&\textcolor{red}{\textbf{2.31}}&\textcolor{red}{\textbf{90.66}} & \textcolor{red}{\textbf{83.37}} & 3.61 & 0.93\\
				\hline
				\multicolumn{1}{|c|}{\multirow{2}*{Dataset}}& \multicolumn{2}{c|}{\multirow{2}*{Scans Used}} &\multicolumn{4}{c|}{LA Dataset}& \multicolumn{4}{c|}{NIH-Pancreas Dataset}& \multicolumn{4}{c|}{ACDC Dataset}\\\cline{4-15}
				{}&{} &{} &\multicolumn{4}{c|}{Metrics}&\multicolumn{4}{c|}{Metrics}&\multicolumn{4}{c|}{Metrics}  \\\cline{1-15}
				\multicolumn{1}{|c|}{\multirow{1}*{Model}}&{Labeled} & {Unlabled} & Dice$\uparrow$ & Jaccard$\uparrow$ & 95HD$\downarrow$ & ASD$\downarrow$& Dice$\uparrow$ & Jaccard$\uparrow$ & 95HD$\downarrow$ & ASD$\downarrow$& Dice$\uparrow$ & Jaccard$\uparrow$ & 95HD$\downarrow$ & ASD$\downarrow$\\ \cline{1-15}
				UA-MT& \multirow{9}*{(10\%)} &\multirow{9}*{(90\%)}& 87.79 & 78.39 & 8.68 & 2.12&66.96&51.89&21.65&6.25& 81.65 & 70.64 & 6.88 & 2.02\\
				SASSNet&&& 87.54 & 78.05 & 9.84 & 2.59&66.69&51.66&18.88&5.76& 84.50 & 74.34 & 5.42 & 1.86 \\
				DTC&&& 87.51 & 78.17 & 8.23 & 2.36&67.28&52.86&17.74&1.97&84.29 & 73.92 & 12.81 & 4.01 \\
				URPC&&&86.92 & 77.03 & 11.13 & 2.28&64.73&49.62&21.90&7.73&83.10 & 72.41 & 4.84 & 1.53 \\
				MC-Net&&& 87.62 & 78.25 & 10.03 & 1.82&69.07&54.36&14.53&2.28&86.44 & 77.04 & 5.50 & 1.84\\
				SS-Net&&&88.55 & 79.62 & 7.49 & 1.90&67.40&53.06&20.15&3.47&86.78 & 77.62 & 6.07 & 1.40\\
				Co-BioNet&&&89.20&80.68&6.44&1.90&82.49&67.88&6.51&3.26&87.66&78.61& \textcolor{red}{\textbf{1.98}}&0.70 \\
				BCP&&&89.62 & 81.31 & 6.81& 1.76&81.54&69.29&12.21&3.80& 88.84& 80.62 & 3.98 & 1.17 \\
				Ours&&&\textcolor{red}{\textbf{91.99}} & \textcolor{red}{\textbf{84.22}} & \textcolor{red}{\textbf{4.54}} & \textcolor{red}{\textbf{1.25}}&\textcolor{red}{\textbf{83.60}}&\textcolor{red}{\textbf{70.81}}&\textcolor{red}{\textbf{5.81}}&\textcolor{red}{\textbf{1.95}}&\textcolor{red}{\textbf{91.17}} & \textcolor{red}{\textbf{83.74}} &2.36 & \textcolor{red}{\textbf{0.70}}\\
				\hline
				\multicolumn{1}{|c|}{\multirow{2}*{Dataset}}& \multicolumn{2}{c|}{\multirow{2}*{Scans Used}} &\multicolumn{4}{c|}{LA Dataset}& \multicolumn{4}{c|}{NIH-Pancreas Dataset}& \multicolumn{4}{c|}{ACDC Dataset}\\\cline{4-15}
				{}&{} &{}&\multicolumn{4}{c|}{Metrics}&\multicolumn{4}{c|}{Metrics}&\multicolumn{4}{c|}{Metrics}  \\\cline{1-15}
				\multicolumn{1}{|c|}{\multirow{1}*{Model}}&{Labeled} & {Unlabled} & Dice$\uparrow$ & Jaccard$\uparrow$ & 95HD$\downarrow$ & ASD$\downarrow$& Dice$\uparrow$ & Jaccard$\uparrow$ & 95HD$\downarrow$ & ASD$\downarrow$& Dice$\uparrow$ & Jaccard$\uparrow$ & 95HD$\downarrow$ & ASD$\downarrow$\\ \cline{1-15}
				UA-MT& \multirow{9}*{(20\%)} &\multirow{9}*{(80\%)}& 88.88& 80.21&7.32&2.26& 77.26 & 63.82 & 11.90 & 3.06&85.16&75.49&5.91&1.79\\
				SASSNet&&&89.54&81.24&8.24&1.99&  77.66 & 64.08 & 10.93 & 3.05&86.45&77.20&6.63&1.98 \\
				DTC&&&89.42&80.98&7.32&2.10&  78.27 & 64.75 & 8.36 & 2.25&87.10&78.15&6.76&1.99 \\
				URPC&&&88.43&81.15&8.21&2.35&79.09&65.99&11.68&3.31&85.44&76.36&5.93&1.70 \\
				MC-Net&&&90.12&82.12&11.28&2.30&78.17&65.22&6.90&1.55&87.83&79.14&4.94&1.52\\
				SS-Net&&&89.25&81.62&6.45&1.80&79.74&65.42&12.44&2.69&88.32 & 79.64 &2.73 & 0.70\\
				Co-BioNet&&&91.26&83.99&5.17&1.64&84.01&70.00&5.35&2.75&91.01 & 82.91 &3.88 & 1.16\\
				BCP&&&90.34&82.50&6.75&1.77&  82.91 & 70.97& 6.43& 2.25&89.52&81.62&3.69&1.03 \\
				Ours&&&\textcolor{red}{\textbf{92.16}} & \textcolor{red}{\textbf{84.51}} & \textcolor{red}{\textbf{4.69}} & \textcolor{red}{\textbf{1.18}}&\textcolor{red}{\textbf{85.12}} & \textcolor{red}{\textbf{71.37}} & \textcolor{red}{\textbf{4.87}} & \textcolor{red}{\textbf{1.21}}&\textcolor{red}{\textbf{91.57}} & \textcolor{red}{\textbf{84.83}} & \textcolor{red}{\textbf{2.20}} & \textcolor{red}{\textbf{0.68}}\\
				\hline
		\end{tabular}}
	\end{threeparttable}
	\label{1}
\end{table*}

\begin{table*}[h]\footnotesize
	\centering
	\caption{Comparison with state-of-the-art models on BraTS dataset in labeled ratio 5$\%$, 10$\%$ and 20$\%$}
	\renewcommand\arraystretch{1}
	\setlength{\tabcolsep}{3mm}{\begin{tabular}{|c|cc|cccccccc|}
			\hline
			\multirow{2}*{Method} & \multicolumn{2}{c|}{No. of scans used} & \multicolumn{2}{c}{Average}& \multicolumn{2}{c}{WT}&\multicolumn{2}{c}{ET}&\multicolumn{2}{c|}{TC} \\
			\cline{2-11}
			& Labeled & Unlabeled & HD $\downarrow$ & DSC $\uparrow$ & HD $\downarrow$ & DSC $\uparrow$ & HD $\downarrow$ & DSC $\uparrow$ & HD $\downarrow$ & DSC $\uparrow$ \\\hline
			UNet&\multirow{7}*{387 (100\%)} & \multirow{7}*{0 (0\%)} & 10.19& 66.40& 9.21& 76.60& 11.12& 56.10& 10.24& 66.50\\
			nnUNet& &  &4.60& 81.90& 3.64& 91.90& 4.06& 80.97& 4.91& 85.35\\
			UNETR& &  &8.82& 71.10& 8.27& 78.90& 9.35& 58.50 &8.85& 76.10\\
			VNet& &  &5.94& 79.09& 7.05& 87.79& 5.37& 72.84 &5.40& 76.63\\
			VT-Unet& &  &3.43& 87.10& 3.51& 91.90& 2.68& 82.20& 4.10& 87.20\\
			MC-Net+& &  &6.26 &82.01& 7.07& 89.38& 5.17& 76.48 &6.53 &80.18\\
			Co-BioNet&  && 6.51 &82.26& 4.98& 90.46& 8.26& 76.58& 6.27& 79.74\\\cline{1-3}
			MC-Net+& \multirow{3}*{4 (1\%)} &\multirow{3}*{383 (99\%)}& 26.65& 69.05&30.10& 75.42& 20.39 &69.04& 29.46& 62.70\\
			Co-BioNet&  & &14.84& 76.03 &20.29 &81.21& 8.54 &73.27& 15.69& 73.63\\
			Ours & &&\textcolor{red}{\textbf{12.13}}&\textcolor{red}{\textbf{77.20}}&\textcolor{red}{\textbf{17.06}}&\textcolor{red}{\textbf{82.98}}&\textcolor{red}{\textbf{6.61}}&\textcolor{red}{\textbf{73.36}}&\textcolor{red}{\textbf{12.73}}&\textcolor{red}{\textbf{75.25}}\\\cline{1-3}
			MC-Net+ &\multirow{3}*{19 (5\%)} &\multirow{3}*{368 (95\%)}& 8.61 &75.80& 9.15& 79.53 &5.46& 72.76 &11.20 &75.11\\
			Co-BioNet& & & 9.06& 76.82& 12.28& 83.65& 5.89 &72.15& 9.03& 74.66\\
			Ours & &&\textcolor{red}{\textbf{7.29}}&\textcolor{red}{\textbf{79.27}}&\textcolor{red}{\textbf{8.67}}&\textcolor{red}{\textbf{86.77}}&\textcolor{red}{\textbf{4.93}}&\textcolor{red}{\textbf{73.84}}&\textcolor{red}{\textbf{8.26}}&\textcolor{red}{\textbf{77.20}}\\\cline{1-3}
			MC-Net+& \multirow{3}*{39 (10\%)}& \multirow{3}*{348 (90\%)}& 9.96& 79.41& 7.79& 85.23& 8.29& 74.71& 13.81& 78.30\\
			Co-BioNet& &  &7.59& 80.75& 8.77& 86.74& 6.24 &76.13& 7.76& 79.38\\
			Ours & &&\textcolor{red}{\textbf{6.31}}&\textcolor{red}{\textbf{82.77}}&\textcolor{red}{\textbf{7.14}}&\textcolor{red}{\textbf{88.28}}&\textcolor{red}{\textbf{5.26}}&\textcolor{red}{\textbf{78.42}}&\textcolor{red}{\textbf{6.53}}&\textcolor{red}{\textbf{81.60}}\\\cline{1-3}
			MC-Net+& \multirow{3}*{77 (20\%)} &\multirow{3}*{310 (80\%)}& 6.54& 80.42& 7.29&88.55& 6.55& 73.52& 5.78& 79.18\\
			Co-BioNet& & & 6.87 &80.86& 9.36 &88.44& 3.39 &75.85& 7.85& 78.28\\
			Ours & &&\textcolor{red}{\textbf{5.31}}&\textcolor{red}{\textbf{83.69}}&\textcolor{red}{\textbf{6.86}}&\textcolor{red}{\textbf{91.14}}&\textcolor{red}{\textbf{2.95}}&\textcolor{red}{\textbf{78.26}}&\textcolor{red}{\textbf{6.13}}&\textcolor{red}{\textbf{81.67}}\\
			\cdashline{1-11}[2pt/2pt]
			\multirow{2}*{Method$^{*}$} & \multicolumn{2}{c|}{No. of scans used} & \multicolumn{2}{c}{\multirow{2}*{Dice$\uparrow$}}& \multicolumn{2}{c}{\multirow{2}*{Jaccard$\uparrow$}}&\multicolumn{2}{c}{\multirow{2}*{95HD$\downarrow$}}&\multicolumn{2}{c|}{\multirow{2}*{ASD$\downarrow$}} \\
			\cline{2-3}
			& Labeled & Unlabeled &&&&&&&& \\\hline
			V-Net&19 (5\%)&368 (95\%)&\multicolumn{2}{c}{76.73}&\multicolumn{2}{c}{65.31}&\multicolumn{2}{c}{7.29}&\multicolumn{2}{c|}{1.45}\\
			V-Net&39 (10\%)&348 (90\%)&\multicolumn{2}{c}{79.79}&\multicolumn{2}{c}{69.56}&\multicolumn{2}{c}{6.18}&\multicolumn{2}{c|}{0.75}\\
			V-Net &387 (100\%)&0 (0\%)&\multicolumn{2}{c}{\textbf{82.54}}&\multicolumn{2}{c}{\textbf{74.55}}&\multicolumn{2}{c}{\textbf{5.08}}&\multicolumn{2}{c|}{\textbf{0.53}}\\\cline{1-3}
			Co-BioNet&\multirow{2}*{19 (5\%)}&\multirow{2}*{368 (95\%)}&\multicolumn{2}{c}{77.20}&\multicolumn{2}{c}{67.56}&\multicolumn{2}{c}{7.81}&\multicolumn{2}{c|}{0.97}\\
			Ours&& & \multicolumn{2}{c}{\textcolor{red}{\textbf{81.96}}}&\multicolumn{2}{c}{\textcolor{red}{\textbf{70.19}}}&\multicolumn{2}{c}{\textcolor{red}{\textbf{5.14}}}&\multicolumn{2}{c|}{\textcolor{red}{\textbf{0.92}}} \\\cline{1-3}
			Co-BioNet&\multirow{2}*{39 (10\%)}&\multirow{2}*{348 (90\%)}&\multicolumn{2}{c}{80.87}&\multicolumn{2}{c}{71.03}&\multicolumn{2}{c}{6.94}&\multicolumn{2}{c|}{0.48}\\
			Ours&& &\multicolumn{2}{c}{\textcolor{red}{\textbf{83.63}}}&\multicolumn{2}{c}{\textcolor{red}{\textbf{73.28}}}&\multicolumn{2}{c}{\textcolor{red}{\textbf{5.46}}}&\multicolumn{2}{c|}{\textcolor{red}{\textbf{0.37}}} \\\cline{1-3}
			Co-BioNet&\multirow{2}*{387 (100\%)}&\multirow{2}*{0 (0\%)}&\multicolumn{2}{c}{82.53}&\multicolumn{2}{c}{72.94}&\multicolumn{2}{c}{5.23}&\multicolumn{2}{c|}{0.44}\\
			Ours&& &\multicolumn{2}{c}{\textcolor{red}{\textbf{84.54}}}&\multicolumn{2}{c}{\textcolor{red}{\textbf{75.66}}}&\multicolumn{2}{c}{\textcolor{red}{\textbf{4.33}}}&\multicolumn{2}{c|}{\textcolor{red}{\textbf{0.33}}} \\\hline
	\end{tabular}}
	\label{THUMOS}%
\end{table*}

\section{Experiments}\label{sec:exp}

\subsection{Experiments Settings}
\begin{figure*}
	\centering 
	\includegraphics[width = 18cm]{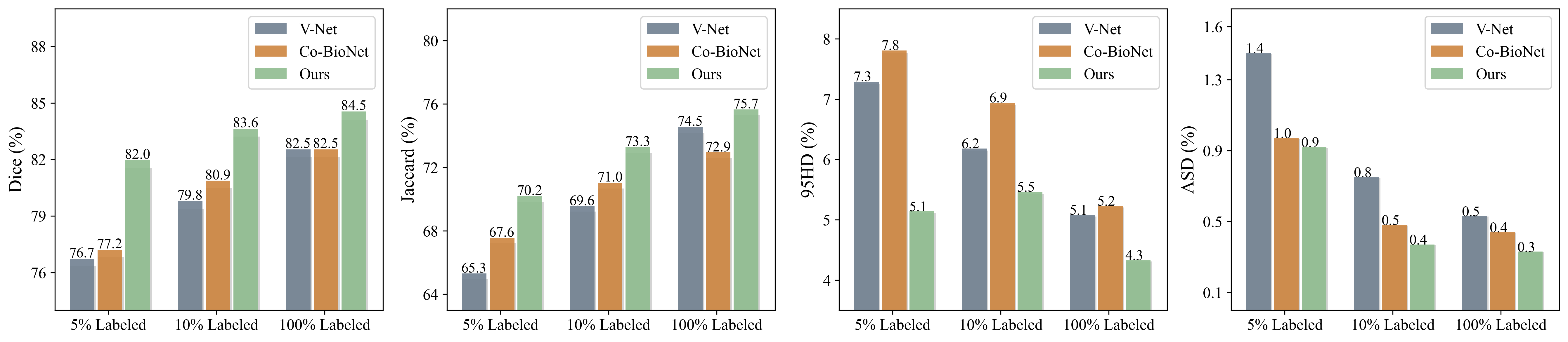}
	\caption{Comparative Results of Four Metrics on BraTS Dataset}
	\label{fgfgfg}
\end{figure*}
\subsubsection{Datasets and Implementation Details} Four public benchmark datasets are employed to evaluate the performance of our proposed model, including the National Institutes of Health (NIH) Pancreas CT dataset, the Left Atrial Segmentation Challenge (LA) dataset, Automated Cardiac Diagnosis Challenge (ACDC) dataset and Brain Tumor Segmentation (BraTS) dataset. The performance is assessed by four metrics:  Dice Score (\%), Jaccard Score (\%), 95\% Hausdorff Distance (95HD) in voxel and Average Surface Distance (ASD) in voxel. The former two metrics primarily compute the percentage of overlap between two object regions. For the latter two metrics, the closest point distance between them is measured by 95HD, while the average distance between them is calculated by ASD (Each batch contains an equal number of labeled and unlabeled samples. Unless otherwise specified, the number of labeled and unlabeled samples in each batch is half of the total batch size. The settings for iterations and epochs can be referenced from BCP \cite{DBLP:conf/cvpr/BaiCL0023}).

\textbf{LA dataset.} LA \cite{xiong2021global} dataset contains a total number of 154 3D magnetic resonance image scans (MRIs) from patients with the $0.625\times0.625\times0.625$ mm isotropic resolution. Each 3D MRI patient data comprises the raw MRI scan and corresponding ground truth labels which are obtained by manual segmentation for the left atrial (LA) cavity. Besides, 100 patient data is used for training, and 54 patient data will be used for testing and evaluation in the dataset. Due to the deficiency of labels in the testing set, we employ 80 samples for training and 20 samples for validation in the training set. For the experimental setup on this dataset, we adopt the configuration used in SS-Net \cite{DBLP:conf/miccai/WuWWGC22}, which employs a 3D V-net backbone and the SGD optimizer. The model is trained with an initial learning rate of 0.01, decaying by 10\% every 2.5K iterations. The pre-training and self-training phases are set to 2K and 15K iterations, respectively. Moreover, the 3D MRIs will be randomly cropped into patches with the given region-of-interest (RoI) size $112\times112\times80$ and the size of the zero-value region of mask $\mathcal{M}_{1,2}$ is set to $74\times74\times53$. And each batch has the same number of labeled and unlabled samples, whose size is set to 4, respectively. \\
\begin{figure}
	\centering 
	\includegraphics[width = 18cm]{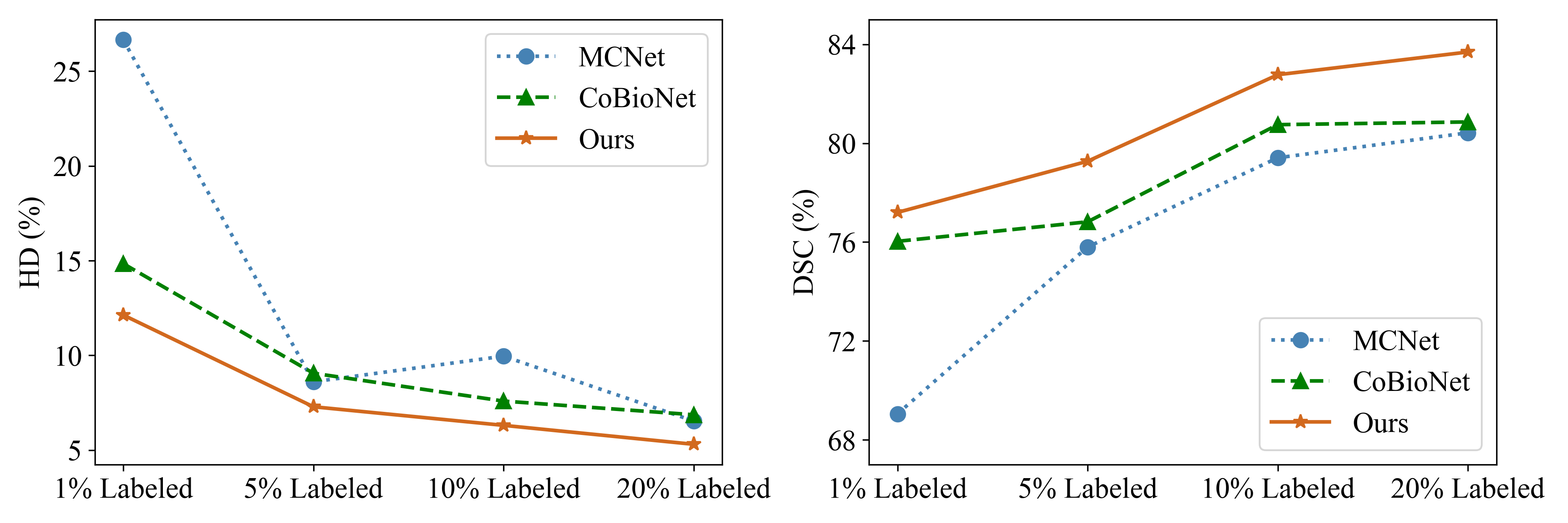}
	\caption{Average Metrics on BraTS Dataset: Comparative Results}
	\label{rtrtrt}
\end{figure}
\begin{figure*}
	\centering 
	\includegraphics[scale=0.25]{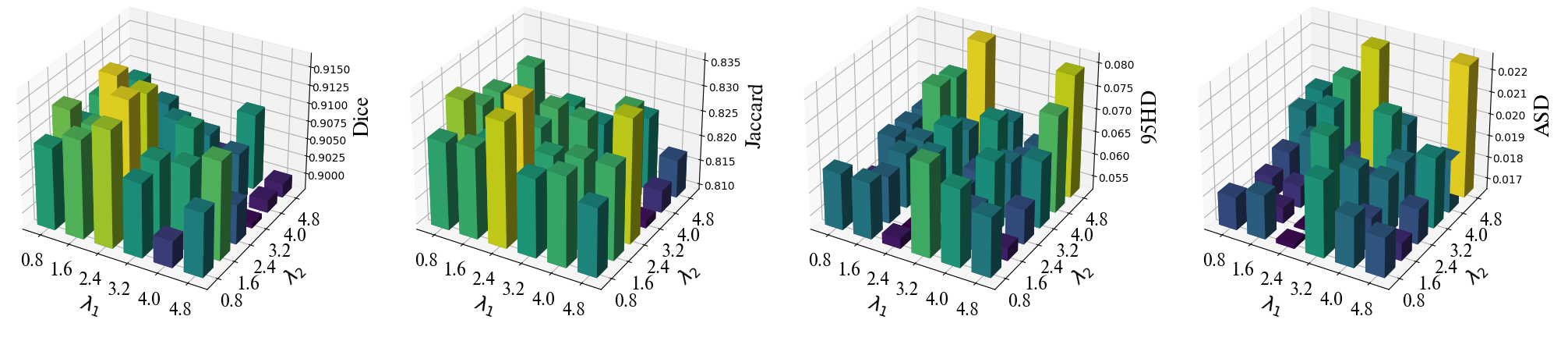}
	\caption{Performance Comparisons with Variation of Parameter $\lambda_1$ and $\lambda_2$}
	\label{lambda1}
\end{figure*}
\begin{figure*}
	\centering 
	\includegraphics[scale=0.25]{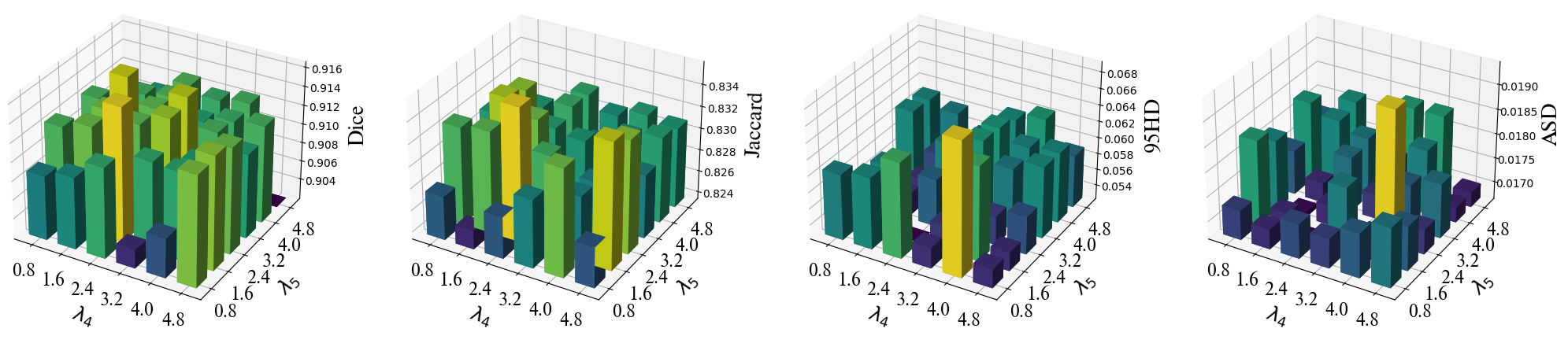}
	\caption{Performance Comparisons with Variation of Parameter $\lambda_4$ and $\lambda_5$}
	\label{lambda2}
\end{figure*}
\begin{figure*}
	\centering 
	\includegraphics[scale=0.25]{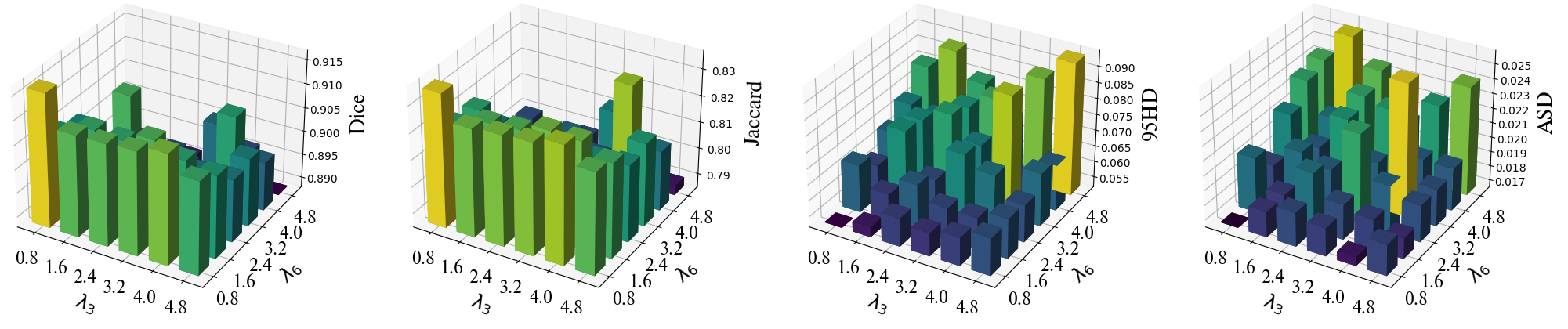}
	\caption{Performance Comparisons with Variation of Parameter $\lambda_3$ and $\lambda_6$}
	\label{lambda3}
\end{figure*}
\textbf{NIH Pancreas-CT}: NIH Pancreas-CT \cite{roth2015deeporgan} is a set of abdominal enhanced CT scan image that is obtained by Siemens multi-detector computed tomography scanners (MDTC), which includes data from 53 male and 27 female subjects. The parameters of the CT scanner are 512 $\ast$ 512 pixels, and the slice thickness ranges from 1.5 to 2.5 mm. Following the previous studies \cite{DBLP:journals/tmi/Shi0LL0Y0022}, 18 samples of the training set are split for validation and the remaining samples are used for training. In addition, the Pancreas CT scans are also cropped into patches with the special RoI size of $96\times96\times96$ and the size of the zero-value region of mask $\mathcal{M}_{1,2}$ is $64\times64\times64$. We adopt an experimental setup similar to CoraNet \cite{DBLP:journals/tmi/Shi0LL0Y0022}, which uses a four-layer 3D V-net as the backbone. The model is trained with the Adam optimizer and an initial learning rate of 0.001. Each batch contains an equal number of labeled and unlabeled samples, with the batch size uniformly set to 4. The pre-training phase consists of 60 epochs, followed by 200 epochs for self-training. Besides, some conventional operations like rotating, rescaling and flipping are exploited to augment data.
\begin{table*}[htbp]\scriptsize
	\renewcommand{\arraystretch}{1.2}
	\caption{Ablation Study of balancing parameter $\lambda$s on LA Dataset with $5\%$ Labeled Ratios}
	\centering\setlength{\belowcaptionskip}{-0.5cm}
	\setlength{\tabcolsep}{1.2mm}{\begin{tabular}{|cc|cccc|cc|cccc|cc|cccc|}
			\hline
			\multicolumn{2}{|c|}{\multirow{2}*{$\lambda$}} &\multicolumn{4}{c|}{LA Dataset} &\multicolumn{2}{c|}{\multirow{2}*{$\lambda$}} &\multicolumn{4}{c|}{LA Dataset}&\multicolumn{2}{c|}{\multirow{2}*{$\lambda$}} &\multicolumn{4}{c|}{LA Dataset} \\
			\multicolumn{2}{|c|}{{}}&\multicolumn{4}{c|}{Metrics (Labeled Ratio $5\%$)}&\multicolumn{2}{c|}{{}}&\multicolumn{4}{c|}{Metrics (Labeled Ratio $5\%$)}&\multicolumn{2}{c|}{{}}&\multicolumn{4}{c|}{Metrics (Labeled Ratio $5\%$)}\\
			\hline
			\multicolumn{2}{|c|}{{$\lambda_1=0.8$}}& Dice$\uparrow$ & Jaccard$\uparrow$ & 95HD$\downarrow$ & ASD$\downarrow$&\multicolumn{2}{c|}{{$\lambda_1=1.6$}}&Dice$\uparrow$ & Jaccard$\uparrow$ & 95HD$\downarrow$ & ASD$\downarrow$&\multicolumn{2}{c|}{{$\lambda_1=2.4$}}&Dice$\uparrow$ & Jaccard$\uparrow$ & 95HD$\downarrow$ & ASD$\downarrow$ \\
			\hline
			\multirow{6}*{$\lambda_2=$}&0.8 & 90.95& 82.69 & 6.49 &1.81 & \multirow{6}*{$\lambda_2=$}&0.8& 91.27& 83.22 &5.71 &1.66 &\multirow{6}*{$\lambda_2=$}&0.8& 90.87& 82.84 & 5.70 &1.76  \\
			&1.6&91.18 & 82.75 & 6.49 &1.86 & &1.6& 91.14  & 82.59 & 6.29 & 1.73&&1.6& 90.65  & 82.38 &6.46  &1.76   \\
			&2.4 & 91.43 & 83.42 &5.48  &1.68 &&2.4 & \textbf{91.63} &\textbf{83.58} & \textbf{5.26} &\textbf{1.67}&&2.4 &90.52  &82.73 & 6.61 &2.03 \\
			&3.2 & 90.84& 82.51 &7.29&2.01 & &3.2& 90.93  & 82.65 & 5.32 &1.74& &3.2& 90.41  & 81.84 & 6.05 &1.92  \\
			&4.0 & 90.18 & 82.74 & 6.95 &1.89 &&4.0 & 90.97 &82.76 &6.18  &1.81&&4.0 & 90.32 & 81.76& 6.86 & 1.91\\
			&4.8 & 90.70 & 82.32 & 6.56 &1.84 &&4.8 & 91.17 &82.78 & 5.54 &1.74&&4.8 & 91.36 &83.41 & 6.02 &1.82 \\
			\hline
			\multicolumn{2}{|c|}{{$\lambda_1=3.2$}}& Dice$\uparrow$ & Jaccard$\uparrow$ & 95HD$\downarrow$ & ASD$\downarrow$&\multicolumn{2}{c|}{{$\lambda_1=4.0$}}&Dice$\uparrow$ & Jaccard$\uparrow$ & 95HD$\downarrow$ & ASD$\downarrow$&\multicolumn{2}{c|}{{$\lambda_1=4.8$}}&Dice$\uparrow$ & Jaccard$\uparrow$ & 95HD$\downarrow$ & ASD$\downarrow$ \\
			\hline
			\multirow{6}*{$\lambda_2=$}&0.8 & 91.04& 82.83 & 6.32 &1.81 & \multirow{6}*{$\lambda_2=$}&0.8&90.54 & 82.31 &6.30 &1.94 &\multirow{6}*{$\lambda_2=$}&0.8& 90.89 & 82.78& 6.28 &1.97  \\
			&1.6&  90.63 & 82.36  & 6.43& 1.91&&1.6& 90.11  & 81.46 & 7.31 & 1.99&&1.6& 90.62  &81.61  & 7.22 & 2.06  \\
			&2.4 & 90.08 & 82.82 & 6.89 & 1.78&&2.4 & 90.75 &82.59 & 6.53 &1.81&&2.4 & 89.82 &80.95 & 8.14 &2.23 \\
			&3.2 &90.96 & 82.76 & 6.11 &1.80 & &3.2& 90.67  & 82.39 & 6.92 &2.03& &3.2&  90.18 & 81.81 & 6.58 &1.92  \\
			&4.0 & 89.81 & 81.49 & 6.50 &1.92 &&4.0 & 90.51 & 82.85& 6.24 &1.83&&4.0 & 90.86 & 82.39& 6.20 & 1.74\\
			&4.8 & 89.88 & 81.10 & 6.72 &1.98 &&4.8 & 89.98 &81.39 &7.38  &1.90&&4.8 & 90.00 & 81.71& 7.94 &2.26 \\
			
			\hline
			\multicolumn{2}{|c|}{{$\lambda_4=0.8$}}& Dice$\uparrow$ & Jaccard$\uparrow$ & 95HD$\downarrow$ & ASD$\downarrow$&\multicolumn{2}{c|}{{$\lambda_4=1.6$}}&Dice$\uparrow$ & Jaccard$\uparrow$ & 95HD$\downarrow$ & ASD$\downarrow$&\multicolumn{2}{c|}{{$\lambda_4=2.4$}}&Dice$\uparrow$ & Jaccard$\uparrow$ & 95HD$\downarrow$ & ASD$\downarrow$ \\
			\hline
			\multirow{6}*{$\lambda_5=$}&0.8 &90.89 & 82.76 & 6.06 &1.73 & \multirow{6}*{$\lambda_5=$}&0.8&91.25 & 83.25 &5.47 &1.84 &\multirow{6}*{$\lambda_5=$}&0.8& 90.99& 82.87 &5.63  &1.81  \\
			&1.6&  90.97 & 82.53 & 6.13& 1.71&&1.6& 91.34  & 83.29 & 6.09 & 1.71&&1.6& 91.36  &83.46  & 5.53 &  1.68 \\
			&2.4 & 91.17 & 82.74 & 6.42 & 1.74&&2.4 & \textbf{91.63} &\textbf{83.58} & \textbf{5.26} &\textbf{1.67}&&2.4 & 91.30 & 83.34& 5.84 &1.71 \\
			&3.2 &90.41 & 82.98 &5.51  &1.73 & &3.2&  91.16 & 83.23 & 5.57 &1.80& &3.2& 91.45  & 82.56 & 5.71 & 1.67 \\
			&4.0 & 90.63 & 83.34 & 6.89 & 1.76&&4.0 & 91.14 &82.95 & 6.41 &1.72&&4.0 & 91.08 & 82.93& 5.61 & 1.94\\
			&4.8 & 91.38 & 82.73 & 5.49 & 1.79&&4.8 & 91.41 &83.51 & 5.49 &1.76&&4.8 & 91.33 & 83.39& 5.72 & 1.72\\
			\hline
			\multicolumn{2}{|c|}{{$\lambda_4=3.2$}}& Dice$\uparrow$ & Jaccard$\uparrow$ & 95HD$\downarrow$ & ASD$\downarrow$&\multicolumn{2}{c|}{{$\lambda_4=4.0$}}&Dice$\uparrow$ & Jaccard$\uparrow$ & 95HD$\downarrow$ & ASD$\downarrow$&\multicolumn{2}{c|}{{$\lambda_4=4.8$}}&Dice$\uparrow$ & Jaccard$\uparrow$ & 95HD$\downarrow$ & ASD$\downarrow$ \\
			\hline
			\multirow{6}*{$\lambda_5=$}&0.8 & 91.25&  83.11& 5.64 &1.76 & \multirow{6}*{$\lambda_5=$}&0.8& 90.75& 82.36 &6.12 &1.83 &\multirow{6}*{$\lambda_5=$}&0.8&90.95 & 82.71 & 6.10 & 1.77 \\
			&1.6&  91.56 & 83.40 &5.54 &1.71 &&1.6&  91.23 & 83.12 & 5.67 &1.81 &&1.6&  91.08 & 82.88 & 6.02 &  1.82 \\
			&2.4 & 91.31 & 83.13 & 6.09 &1.78 &&2.4 & 91.28 & 83.20& 5.91 &1.79&&2.4 & 91.27 & 83.18& 5.81 & 1.74\\
			&3.2 & 91.52&83.09  & 6.24 &1.72 & &3.2&  91.11 & 82.94 & 6.22 &1.76& &3.2& 91.19  &83.06  &6.14  & 1.84 \\
			&4.0 & 91.24 & 82.64 & 6.00 &1.76 &&4.0 &91.20  &83.07 & 6.08 &1.80&&4.0 & 91.24 & 83.14& 6.25 & 1.84\\
			&4.8 & 91.10 & 82.94 & 6.13 & 1.78&&4.8 & 91.24 &83.14 & 6.12 &1.70&&4.8 & 91.21 &83.09 & 5.90 &1.70 \\
			
			\hline
			\multicolumn{2}{|c|}{{$\lambda_3=0.8$}}& Dice$\uparrow$ & Jaccard$\uparrow$ & 95HD$\downarrow$ & ASD$\downarrow$&\multicolumn{2}{c|}{{$\lambda_3=1.6$}}&Dice$\uparrow$ & Jaccard$\uparrow$ & 95HD$\downarrow$ & ASD$\downarrow$&\multicolumn{2}{c|}{{$\lambda_3=2.4$}}&Dice$\uparrow$ & Jaccard$\uparrow$ & 95HD$\downarrow$ & ASD$\downarrow$ \\
			\hline
			\multirow{6}*{$\lambda_6=$}&0.8 & \textbf{91.63} &\textbf{83.58} & \textbf{5.26} &\textbf{1.67}& \multirow{6}*{$\lambda_6=$}&0.8& 90.48& 81.84 &6.74 &2.04 &\multirow{6}*{$\lambda_6=$}&0.8& 90.34& 81.89 &6.26  &1.89  \\
			&1.6&90.96 & 82.62 & 5.57 & 1.84& &1.6& 90.76  & 82.27 & 6.11 & 1.85&&1.6&  89.85 & 90.31 & 7.61 & 2.04  \\
			&2.4 & 90.97 & 82.71 & 6.14 &1.90 &&2.4 & 90.72 & 82.35& 6.69 &2.06&&2.4 & 88.87 &79.83 & 6.36 &1.98 \\
			&3.2 & 91.00& 82.77 & 5.98 &1.86 & &3.2& 90.91  & 82.63 & 6.25 &1.91& &3.2& 89.75  & 80.81 & 7.43 & 2.28 \\
			&4.0 & 91.14 & 83.01 & 6.24 & 1.73&&4.0 & 90.68 & 82.58& 6.23 &1.87&&4.0 & 90.10 &81.36 & 7.07 &1.99 \\
			&4.8 & 90.78 & 82.38 & 6.42 & 1.88&&4.8 & 90.56 &82.12 & 6.50 &1.81&&4.8 & 90.15 &81.46 & 6.42 &1.92 \\
			\hline
			\multicolumn{2}{|c|}{{$\lambda_3=3.2$}}& Dice$\uparrow$ & Jaccard$\uparrow$ & 95HD$\downarrow$ & ASD$\downarrow$&\multicolumn{2}{c|}{{$\lambda_3=4.0$}}&Dice$\uparrow$ & Jaccard$\uparrow$ & 95HD$\downarrow$ & ASD$\downarrow$&\multicolumn{2}{c|}{{$\lambda_3=4.8$}}&Dice$\uparrow$ & Jaccard$\uparrow$ & 95HD$\downarrow$ & ASD$\downarrow$ \\
			\hline
			\multirow{6}*{$\lambda_6=$}&0.8 &89.68 & 80.77 & 6.84 & 2.14& \multirow{6}*{$\lambda_6=$}&0.8& 88.97& 79.49 &7.30 & 2.28&\multirow{6}*{$\lambda_6=$}&0.8 & 89.18 & 89.97 & 8.07&2.33 \\
			&1.6& 90.87& 80.58 & 7.51 & 1.97& &1.6&  89.38 & 80.18 & 7.14 & 2.07&&1.6&  88.91 & 78.59 & 8.81 & 2.54  \\
			&2.4 & 88.87 & 78.79 & 7.94 & 2.22&&2.4 & 89.54 & 80.53& 7.71 &2.28&&2.4 & 89.09 & 79.84& 8.02 & 2.36\\
			&3.2 & 89.75& 80.58 & 7.25 & 1.98& &3.2&  89.28 & 79.98 & 8.22 &2.24& &3.2&  89.99 & 81.12 & 7.10 &2.05  \\
			&4.0 & 89.62 & 80.19 & 9.03 & 2.57&&4.0 & 90.64 &82.98 & 6.02 &1.96&&4.0 & 89.58 & 79.87& 8.67 &2.23 \\
			&4.8 & 90.26 & 81.69 &6.91  &1.92 &&4.8 & 89.86 &80.83 & 6.68 &1.96&&4.8 &88.85  & 78.93& 9.40 &2.41 \\
			
			\hline
	\end{tabular}}
	\label{ablation2}
\end{table*}

\textbf{ACDC dataset.}  ACDC \cite{bernard2018deep} is a heart 3D MRI Image dataset that is obtained by two scanners and annotated with three classes including left ventricle (LV), right ventricle (RV) and myocardium (Myo) in diastolic (ED) and systolic (ES) frames of cardiac dynamic magnetic resonance imaging (cine MRI). It covers 150 patients' MRI scans divided into five identical subgroups. Among them, only 100 samples' annotations are publicly released which are employed for cross-dataset validation. Similarly, we segment these MRI data with annotations into training, validation, and testing sets with a 7:1:2 splitting ratio. Additionally, we train a 2D U-Net using the SGD optimizer with an initial learning rate of 0.01. The input patch size is set to $256 \times 256$, and the zero-value region of the mask has dimensions of $64 \times 64 \times 64$, following the experimental setup in SS-Net \cite{DBLP:conf/miccai/WuWWGC22}. Each batch contains an equal number of labeled and unlabeled samples, with the batch size uniformly set to 12. The pre-training consists of 10K iterations, followed by 30K iterations for self-training.

\textbf{MSD BraTS dataset.} MSD BraTS dataset \cite{antonelli2022medical} is a multi-class and multi-modality brain tumour dataset which is significantly different from NIH Pancreas-CT and LA datasets. As before, we split the training dataset which contains 484 MR volumes with the shape of $240\times240\times155$ into three subsets for training, validation and testing with a 16:3:1 splitting ratio. The MR volumes are selected from four categories of MRI sequences including a T1-weighted sequence (T1), a T1-weighted contrast-enhanced sequence (T1CE) using gadolinium contrast agents (T1Gd), a T2-weighted sequence (T2) and a fluid-attenuated inversion recovery (FLAIR) sequence. The enhancing tumor (ET), which is the area of relative hyper-intensity in the T1CE with respect to the T1 sequence, the necrotic tumor (NCR) and non-enhancing tumor (NET), which are both hypo-intense in T1-Gd when compared to T1, and peritumoral edema (ED), which is hyper-intense in the FLAIR sequence, are the four distinct tumor sub-regions identified by using these sequences. And then, the (almost) homogeneous sub-regions are transformed into three types of tumors that have semantic meaning: (1) enhancing tumour (ET); (2) tumour core (TC) region which is represented by ET, NET, and NCR; and (3) the complete tumour (WT) which is denoted by ED. As for the data pre-process, the MR volumes are cropped into patches with the shape of $128\times128\times96$. And the model is trained using the SGD optimizer with an initial learning rate of 0.01 and a momentum of 0.9. The pre-training and self-training phases are set to 2K and 15K iterations, respectively. And each batch has the same number of labeled and unlabled samples, whose size is uniformly set to 4.

\subsection{Discussions on the Experimental Results on LA, NIH Pancreas-CT and ACDC Dataset}
We conduct the abundant experiments with different labeled ratios ($i.e.$ 5\%, 10\% and 20\%) on LA, NIH Pancreas-CT and ACDC datasets and compare the performance of our proposed method with fully supervised method V-Net/U-net and previous weakly supervised methods, including UA-MT \cite{DBLP:conf/miccai/YuWLFH19}, SASSNet \cite{DBLP:conf/miccai/LiZH20}, DTC \cite{DBLP:conf/aaai/LuoCSW21}, URPC \cite{DBLP:conf/miccai/LuoLCSCZCWZ21}, MC-Net \cite{DBLP:conf/miccai/WuXGCZ21}, SS-Net \cite{DBLP:conf/miccai/WuWWGC22}, Co-BioNet \cite{peiris2023uncertainty}, BCP \cite{DBLP:conf/cvpr/BaiCL0023}. Besides, it should be noted that all of these methods don't carry out full experiments and we complete their missing experiments with specific labeled ratios. Table \ref{1} illustrates the experimental results where the best performance is highlighted in red. As is shown in Table \ref{1}, our method outperforms the most advanced method BCP for three labeled ratios by a huge margin on LA dataset. Specifically, when the ratio of labeled data is 5 \%, our method transcends BCP by 3.61\%, 4.86\%, 2.64 and 0.48 on Dice, Jaccard, 95HD and ASD respectively. In addition, there is an interesting phenomenon that the performance of Co-BioNet is significantly worse when the percentage of label data is 5\% than when it is 10\% or 20\%. For the dice metric, the average effect of the proposed method $(91.92)$ is 2.6 higher than that of the best existing method $(89.32)$ in the three labeling ratios.

For the experimental results on NIH Pancreas-CT dataset, it can be observed that the performance of all methods on the four given metrics is worse than that on LA dataset. Besides, except for Co-BioNet and BCP, the other comparison methods show a tremendous performance gap as the proportion of annotated data increases and demonstrate disadvantageous performance on this dataset, which reveals the intricacy of segmenting the pancreas and the variation among various tasks related to medical picture segmentation. Remarkably, the performance of our method on NIH-Pancreas dataset with 5\% labeled data surpasses that of BCP under the setting of 10\% labeled data. 

On ACDC dataset, our method also outperforms previous SOTA methods, especially on the latter two indicators. In details, the performance of our method exceeds BCP by 0.38 and 0.46 on 95HD and ASD metrics, which is even more dominant than the fully-supervised V-Net. It also indicates that segmentation provided by our method is more accurate and has a cleaner, smoother edge. In sum, the four metrics on ACDC dataset of the proposed method is able to achieve steady performance improvement as the proportion of annotations grows.

\begin{figure*}
	\centering 
	\includegraphics[width = 17cm]{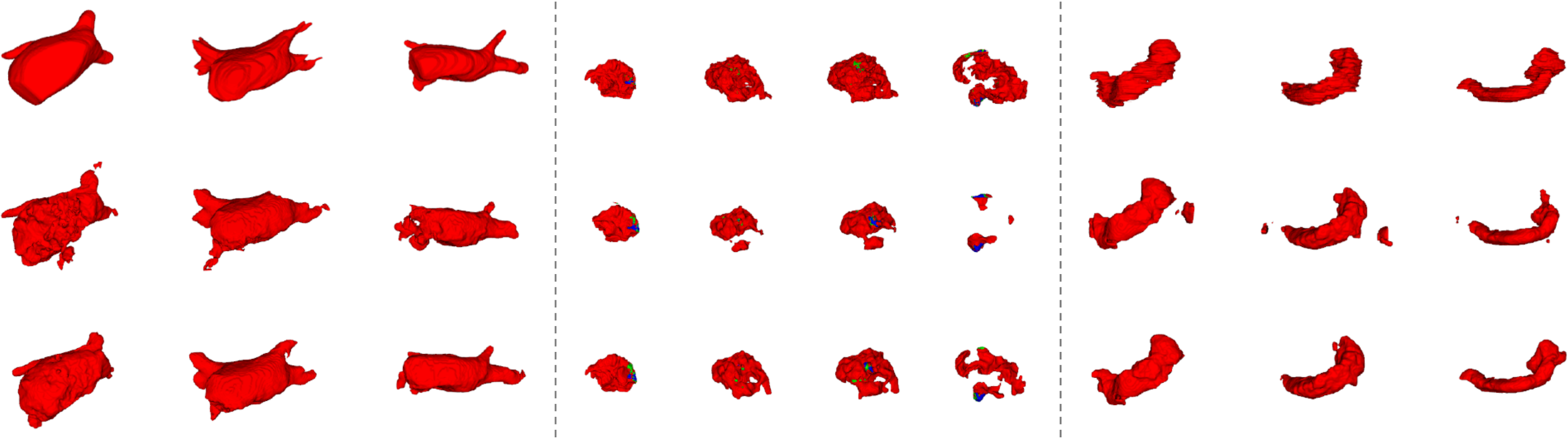}
	\caption{Visualized Comparisons on LA, BraTS and Pancreas dataset. The first row represents the ground truth. In the second row, the first to third columns and the eighth to tenth columns show the segmentation visualization results of BCP, while the fourth to seventh columns display the segmentation visualization results of Co-BioNet. The third row presents the visualization results of the proposed method. The first to third columns correspond to the LA Dataset, the fourth to seventh columns correspond to the BraTS Dataset, and the eighth to tenth columns correspond to the Pancreas Dataset. Each vertical column represents a comparative set of segmentation results for the same slice.}
	\label{11111}
\end{figure*}
\subsection{Discussions on the Experimental Results on BraTS Dataset}
To explore the scalability of our proposed model on massive-scale datasets, we conduct two sections of experiments on the MSD BraTS dataset. For fair comparison, the experiments in the first section of Table \ref{THUMOS} are conducted under the setting of Co-BioNet. It should be noted that the performance of Co-BioNet and other comparison methods is directly extracted from Co-BioNet \cite{peiris2023uncertainty}. Obviously, on three classes (WT, ET and TC), our method achieves optimal performance across three ratios of labeled data on the dataset. To be specific, Co-BioNet and most fully supervised methods are surpassed by our method, which achieves an average performance of 5.31 and 83.69\% on HD and DSC (Dice similarity coefficient) metrics. In the second section of Table \ref{THUMOS}, we also evaluate the property of our method on the dataset with 5\%, 10\% and 100\% labeled data respectively, utilizing the four given metrics. Different from the first section, the performance of V-Net and Co-BioNet is reproduced by ourselves which demonstrates a certain degree of disparity in performance. When the ratio of labeled data is 5\%, the most prominent improvement is achieved on the former three metrics ($i.e.$ transcending Co-BioNet by 4.76\%, 2.63\% and 2.67 respectively). Moreover, when we increase the percentage of labeled data from 10\% to 100\%, there is not an obvious performance improvement which is even lower than that when the proportion of data with labels rises from 5\% to 10\%. It confirms that our method is capable of efficiently extracting assistant knowledge from labeled data to unlabeled data even though the ratio of labeled data is relatively small. 









\subsection{Parameters Study}
In this section, we conduct extensive experiments on LA dataset with 5\% labeled data whose results are presented in Table \ref{ablation2}, to explore the changes in performance as the parameters of aforesaid loss functions change. $\lambda_{1,3}$ and $\lambda_{2,4}$ denote the weights of uncertainty-guided and information volume-related loss functions in the pre-training and self-training stages, respectively. For the first two rows of the table, we adjust the ratios of loss functions $\lambda_{1,2}$ in the pre-training stage. When $\lambda_1 = 1.6$ and $\lambda_2 = 2.4$, our model achieves the best performance (91.63\%, 83.58\%, 5.26, and 1.67 on the four given metrics), which indicates that properly increasing the weight of the loss function related to information volume can better improve the performance of the model. It also verifies the effectiveness of $\textbf{IVUM}$ that makes full use of the concept of information volume in mass function. Similarly, the same optimal ratio of parameters $\lambda_{4,5}$ appears in the loss functions in the self-training stage. Then, $\lambda_{3,6}$ denote the weights of loss function improved by both uncertainty and information volume in the two phases. According to the experimental results in Table \ref{ablation2}, it is rational that when $\lambda_3$ is equal to $\lambda_6$, the best performance is realized because of the same parameter setting of two stages.
\begin{figure}
	\centering 
	\includegraphics[scale=0.3]{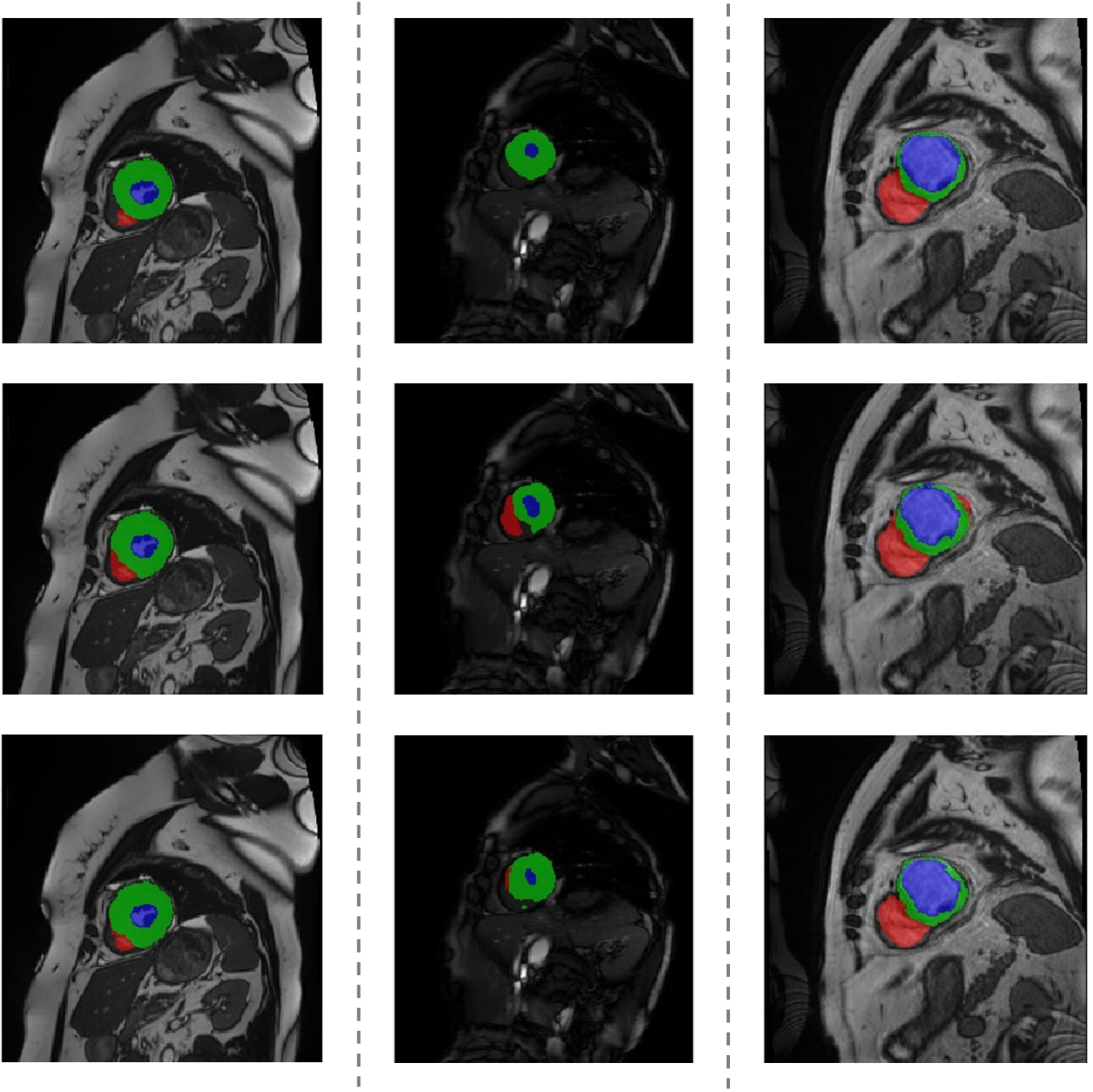}
	\caption{Visualized Comparisons with ground truth and BCP on the ACDC Dataset.The first row represents the ground truth, the second row shows the segmentation visualization results of BCP, and the third row presents the visualization results of the proposed method. Each vertical column corresponds to a comparative set of segmentation results for the same slice.}
	\label{2222222}
\end{figure}
\subsection{Discussions on Visualized Results }
Fig. \ref{11111} demonstrates some typical visualization samples of our method and comparisons on LA, BraTS and Pancreas dataset. It can be observed that our method is capable of providing a smoother and more detailed segmentation compared with other comparison methods, which present common problems like discrete mispredictions and minor edge protrusions or depressions. Especially for the prediction on Pancreas dataset, the segmentation supplied by our method is far superior to others, which possess a higher ratio of overlap with labels and fewer error predictions. In Fig. \ref{2222222}, compared to the state of the art method on the ACDC dataset, BCP, the proposed method can achieve segmentation results that are much closer to the ground truth and have a higher proportion of overlapping regions, which can be easily inferred from the presented images.

\section{Differences with Existing Evidential Deep Learning-based Methods}
Our work fundamentally differs from previous approaches in several key aspects. \textbf{ETCL} \cite{zhang2024evidential} employs evidential fusion for generating pseudo-labels. In contrast, our method introduces the strategy of \textbf{co-evidential fusion} and traditional evidence-theory-based \textbf{generalized evidential learning} into semi-supervised medical image segmentation. The incorporation of Dempster-Shafer (D-S) evidence theory is motivated by the observation that uncertainty and credibility in decision-making play a crucial role in ensuring diagnostic accuracy in medical image analysis. By integrating evidence theory with \textbf{pignistic evidential fusion}, specifically designed for semi-supervised learning, our approach enhances the model's ability to learn from supervisory signals while efficiently managing uncertainty in the relationship between labeled and unlabeled data.  \textbf{EVIL} \cite{chen2023evil} utilizes the original evidential deep learning (\textbf{EDL}) loss and its probability definitions, incorporating them into the Dice loss. In contrast, our approach revisits the relationship between uncertainty and EDL loss by introducing \textbf{information volume of mass function}, a key element in evidence theory, to re-evaluate voxel-level evidence. \textbf{BFSSL} \cite{huang2021belief} combines predictions from UNet and Evidential Neural Networks (\textbf{ENN}) through evidential fusion. Our method, however, further reduces the dependency on large amounts of labeled data by leveraging an \textbf{uncertainty-guided learning paradigm}. Unlike traditional semi-supervised learning approaches, which still exhibit performance limitations when labeled data is scarce, our method constructs meaningful relations between labeled and unlabeled data through uncertainty exploration and evidence evaluation. This enables a more flexible and adaptive utilization of unlabeled data, ultimately improving segmentation performance even in low-data scenarios.  \textbf{EPE} \cite{wang2024evidential} applies evidential fusion for pseudo-label generation using strongly and weakly augmented data. In contrast, our approach employs \textbf{co-evidential fusion} to enhance the relationship between labeled and unlabeled data, ensuring a more structured and reliable transfer of knowledge across different levels of supervision.  \textbf{EUA} \cite{chen2024evidence} proposes combining the conventional EDL loss with a belief-mass-based Dice loss to enhance model performance. While this approach focuses on improving segmentation accuracy, our study extends the EDL framework through \textbf{generalized evidential learning}, incorporating information volume-based evidential optimization objectives under a structured evidential deep learning paradigm, providing distinct contributions beyond the scope of \textbf{EUA}. Overall, the proposed method based on the generalized framework of EDL achieves a more fine-grained voxel-level evaluation through co-evidential fusion and the information volume of the mass function. This approach enables the model to better leverage unlabeled data and more robustly distinguish whether a voxel requires segmentation. It significantly differs from existing methods and achieves state-of-the-art performance.

\section{Conclusion}\label{sec:con}
In this paper, we propose a co-evidential fusion strategy as well as three evidential optimization objectives based on it. The specially designed fusion strategy is used to assist the model in performing fine-grained targeted learning on labeled data and further learning the semantic relationship between labeled and unlabeled data, thereby guiding the model to learn voxel-wise in the pre-training and self-training stages respectively. For the three different evidential optimization objectives, the first one utilizes the optimized uncertainty assessment obtained after the co-evidential fusion to rank the generated relevant evidence accordingly, guiding the model to learn different parts to different extents. The second objective improves the optimization function of traditional evidential deep learning by combining the uncertainty measure after fusion refinement and the information volume of the mass function, thereby designing a new optimization objective. Moreover, the third evidential optimization objective combines the ideas of ranking and evaluating the evidence and the redesigned evidential optimization functions of the first two objectives.

Overall, the proposed method can achieve better results than other existing methods and, in most cases, can achieve better segmentation performance with less labeled data compared to other methods that require more labeled data. However, this might lead to higher computational resource consumption, potentially resulting in a speed disadvantage compared to other methods. At the same time, how to guide the model to carry out targeted learning more efficiently remains an open issue. Future work should focus on better evidence evaluation methods and information fusion strategies in the field of semi-supervised medical image segmentation, which could provide a more comprehensive assessment of finer-grained features and achieve superior segmentation performance.

\section*{Acknowledgment}
This work is supported by National Key R$\&$D Program of China (2023YFC3502900). The authors greatly appreciate the anonymous reviewers’ suggestions and the editor’s encouragement.

\bibliographystyle{elsarticle-num}
\bibliography{cite}

\end{document}